\documentclass[letterpaper]{article} 
\usepackage{aaai23}  
\usepackage{times}  
\usepackage{helvet}  
\usepackage{courier}  
\usepackage[hyphens]{url}  
\usepackage{graphicx} 
\usepackage{natbib}  
\usepackage{caption} 
\usepackage{algorithm}
\usepackage{algorithmic}
\usepackage[table]{xcolor}
\usepackage{dsfont}

\usepackage{bm}
\newcommand{\ie}{\textit{i}.\textit{e}.}
\newcommand{\eg}{\textit{e}.\textit{g}.}
\usepackage{subcaption}
\newcommand{\dbar}{\mathrm{d}}
\usepackage{multirow}
\usepackage{booktabs}
\usepackage{amsfonts} 

\usepackage{amsmath}
\usepackage{amssymb}
\usepackage{mathtools}
\usepackage{amsthm}
\theoremstyle{plain}
\newtheorem{theorem}{Theorem}[]
\newtheorem{proposition}[theorem]{Proposition}

\theoremstyle{definition}

\theoremstyle{remark}

\usepackage{newfloat}
\usepackage{listings}
\DeclareCaptionStyle{ruled}{labelfont=normalfont,labelsep=colon,strut=off} 
\floatstyle{ruled}
\newfloat{listing}{tb}{lst}{}
\floatname{listing}{Listing}
\title{CMVAE: Causal Meta VAE for Unsupervised Meta-Learning}
\author{
Guodong Qi \textsuperscript{\rm 1,2},
Huimin Yu \textsuperscript{\rm 1,2,3,4}\\
}
\affiliations{

\textsuperscript{\rm 1}College of Information Science and Electronic Engineering, Zhejiang University \\ 
\textsuperscript{\rm 2}ZJU-League Research \& Development Center  \text{     }     
\textsuperscript{\rm 3}State Key Lab of CAD\&CG, Zhejiang University \\
\textsuperscript{\rm 4}Zhejiang Provincial Key Laboratory of Information Processing, Communication and Networking \\

\{guodong\_qi, yhm2005\}@zju.edu.cn
}

\usepackage{bibentry}

\begin{document}

\maketitle

\begin{abstract}
	Unsupervised meta-learning aims to learn the meta knowledge from unlabeled data and rapidly adapt to novel tasks. However, existing approaches may be misled by the  context-bias (e.g. background) from the training data. In this paper, we abstract the unsupervised meta-learning problem into a Structural Causal Model (SCM) and point out that such bias arises due to hidden confounders. To eliminate the confounders, we define the priors are \textit{conditionally} independent, learn the relationships between priors and intervene on them with casual factorization. Furthermore, we propose Causal Meta VAE (CMVAE) that encodes the priors into latent codes in the causal space and learns their relationships simultaneously to achieve the downstream few-shot image classification task. Results on toy datasets and three benchmark datasets demonstrate that our method can remove the context-bias and it outperforms other state-of-the-art unsupervised meta-learning algorithms because of bias-removal. Code is available at \url{https://github.com/GuodongQi/CMVAE}.
\end{abstract}

\section{Introduction}

\label{sec:introduction}Regular meta-learning algorithms such as \cite{finn2017model, snellPrototypicalNetworksFewshot2017} aim to learn the meta knowledge to adapt to novel tasks quickly. However, it requires various supervised tasks on large labeled datasets during the meta-training phase. Recently, researchers take great interest in \textit{unsupervised meta-learning} \cite{hsu2018unsupervised, khodadadeh2021unsupervised}. Different from  regular meta-learning, unsupervised meta-learning contains unsupervised meta-training and supervised meta-test. It aims to learn a learning procedure with unlabeled datasets in the meta-training and solve novel supervised human-crafted tasks in the meta-test.

\begin{figure}[]
	\centering
	\begin{subfigure}{.44\columnwidth}
		\centering
		\includegraphics[width=1\columnwidth]{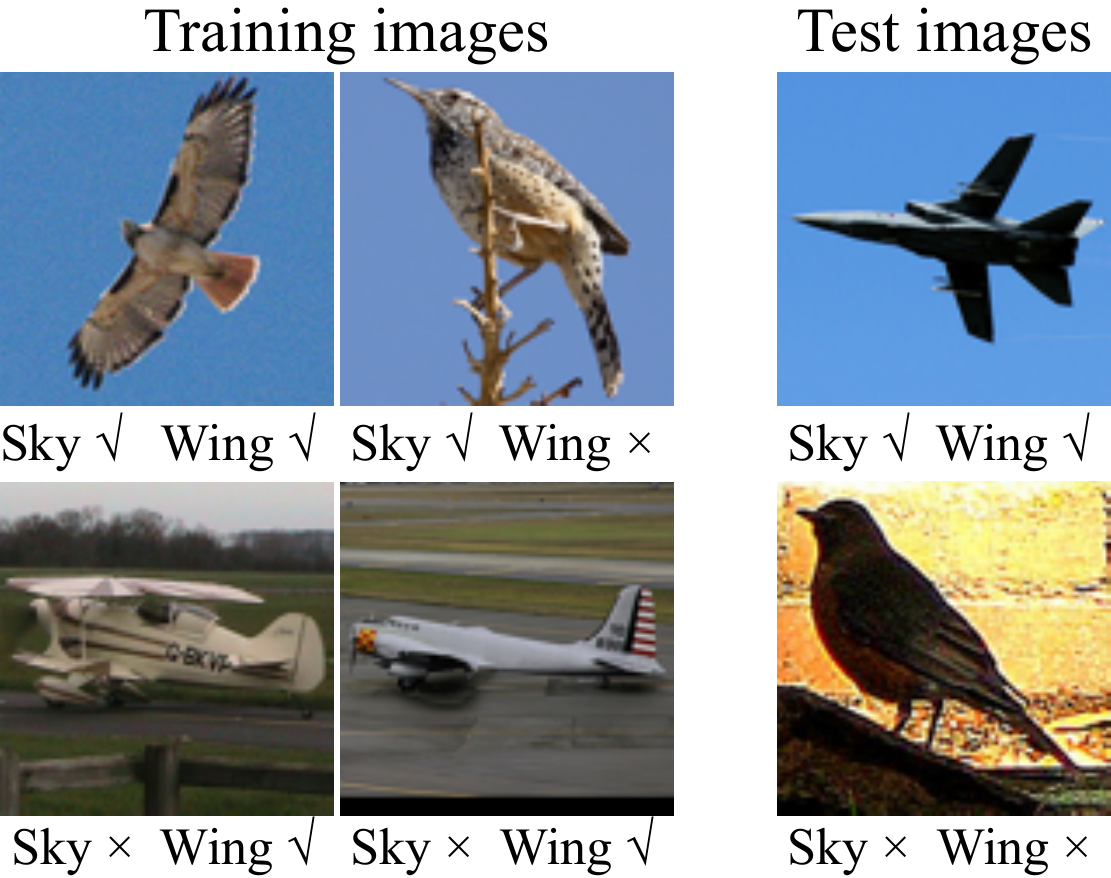}
		\caption{}
		\label{fig:intro}
	\end{subfigure}	
	\begin{subfigure}{.39\columnwidth}
		\centering
		\includegraphics[width=1\columnwidth]{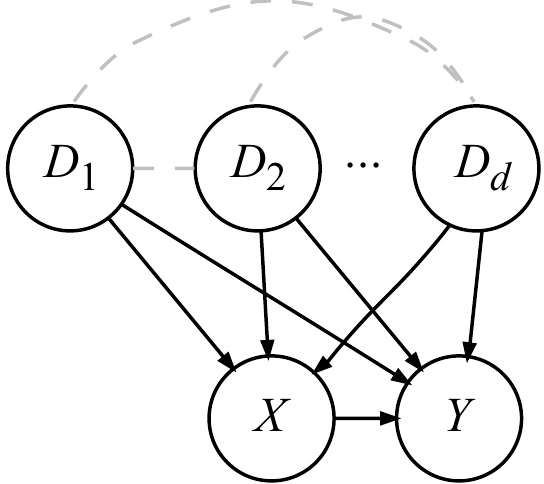}
		\caption{}
		\label{fig:intro_scm}
	\end{subfigure}	
	\caption{(a) Illustration of context-bias. (b) SCM of unsupervised meta-learning. The dashed line means that the relationship (DAG) need to be learned.  }
\end{figure}

Previous methods focus on the pseudo-label generation of the task. However, they may ignore the bias. Figure \ref {fig:intro} illustrates a binary-classification toy example where the background prior is one of bias. In the training images, the ``birds'' are always together with the ``sky'' and the ``airplanes'' always park on the ground. As a result, the model will take the ``sky'' as a part of the ``bird'', and mistakenly recognize the ``airplane'' test image as a ``bird''. It is essential to remove the  effect of background prior \ie, context-bias.

However, discerning the context-bias is challenging, because the priors may not be independent. For example, in the task of Figure \ref{fig:intro}, the ``wing'' and the ``sky'' prior is not independent statistically\footnote{$P(\text{wing}, \text{sky})=1/4$, $P(\text{wing})=3/4$, $P(\text{sky})=1/2$, we have $P(\text{wing}, \text{sky})\neq P(\text{wing})P(\text{sky})$, so they are dependent.}. When the ``sky'' prior is removed, the ``wing'' prior will be changed, and then the prediction will be affected. In this case, the model will not know whether the ``sky'' or ``wing'' prior is the context-bias.

To address the problems, we analyze, discern and remove the context-bias  from a causal perspective via three theories, \ie, Structural Causal Model (SCM) \cite{glymour2016causal}, Common Cause Principle (CCP) \cite{scholkopfCausalRepresentationLearning2021} and Independent Causal Mechanism (ICM) \cite{DBLP:conf/icml/ScholkopfJPSZM12}. Among them, SCM describes the relevant concepts and how they interact with each other. CCP reveals that if two observables are statistically dependent, then there exists a variable such that they are independent conditioned on the variable. ICM states that the conditional distribution of each prior given its causes does not influence the others. In other words, SCM explains how the bias affects predictions.  CCP makes it reasonable to assume the priors are \textit{conditionally} independent.  For example, in Figure \ref{fig:intro} there exists a  ``flying'' prior, which causally affects ``sky'' and ``wing'' and makes them independent when conditioned on the prior. ICM allows us to remove one prior (\eg, $p(\text{sky}|\text{flying})$) will not affect another prior (\eg, $p(\text{wing}|\text{flying})$).

Specially, we build the SCM in Figure \ref{fig:intro_scm}. The bias emerges because the priors are confounders that cause spurious correlations from the inputs to predictions. To achieve bias-removal, we define the relationships between priors based on CCP, obtain the structure with a learnable directed acyclic graph (DAG), causally factorize the joint distribution of priors based on ICM, and then perform causal intervention \cite{glymour2016causal} in sequence.

Furthermore, we design the Causal Meta VAE (CMVAE), which learns the priors and the causal factorization simultaneously. Particularly, we propose the causal intervention formula with the SCM. It leads us to learn the conditionally independent latent codes (priors) as well as the DAG (causal factorization). To make the correspondence between the latent codes and priors, we adopt the VAE-based framework \cite{Kingma2014vae} since VAE has been shown to achieve some useful disentangling performance \cite{higgins2016beta}. The ``DAG-ness'' can be quantified by a regularizer \cite{Zheng2018dags}. Besides, we introduce the Causal Latent Space (CaLS) and show its addability, which makes it feasible to represent the class-concept codes while keeping the DAG. We also extend one baseline \cite{lee2021metagmvae} into our CMVAE to achieve the downstream few-shot classification with the unsupervised meta-learning settings. The contributions of this paper are as follows: 
\begin{itemize}
	\item We point out the context-bias and the dependent priors in unsupervised meta-learning. We propose to learn the relationship among the priors with a learnable DAG and make the priors causally independent and factorize.
	\item We design the intervention formula, introduce the CaLS, and propose CMVAE to learn the factors and the factorization for the downstream classification simultaneously.
	\item Extensive experiments on two toy datasets and three widely used benchmark datasets demonstrate that CMVAE outperforms other state-of-the-art unsupervised meta-learning algorithms. Furthermore, we show that CMVAE can be intervened to generate counterfactual samples with some meaningful explanation.
\end{itemize}

\section{Related Work}

\textbf{Unsupervised Meta-Learning} aims to learn the meta-knowledge with unlabeled training data. CACTU \cite{hsu2018unsupervised} and UMTRA \cite{khodadadeh2018unsupervised} try to create synthetic labels. GMVAE \cite{lee2021metagmvae} introduces a Mixture of Gaussian priors by performing Expectation-Maximization (EM). However, none of them notices the  bias in the few-shot tasks.

\noindent\textbf{Causal Inference} helps machines understand how and why causes influence their effects \cite{glymour2016causal}. Recently, the connection between causality and machine learning \cite{magliacane2018domain, DBLP:conf/iclr/BengioDRKLBGP20, DBLP:conf/nips/KyonoZS20} or computer vision \cite{DBLP:conf/cvpr/Lopez-PazNCSB17, yang2021causal, DBLP:conf/cvpr/WangHZS20a} have gained increasing interest. Recently, IFSL \cite{yue2020inter} introduces the causality into few-shot learning problem with an SCM. However, CMVAE differs since it explicitly learns and utilizes the causal factorization.  

\noindent \textbf{DAG Learning} is to estimate the structure of variables. There are three types of methods, the discrete optimization \cite{scanagatta2016learning, viinikka2020towards}, the continuous optimization \cite{Zheng2018dags, zheng2020learning} and the sampling-based methods \cite{charpentier2021differentiable}. 
CMVAE incorporates recent continuous optimization methods to learn the DAG of the context-priors.

\section{Proposed Formulation}

\subsection{Problem Statement}  

Given an unlabeled dataset $\mathcal{U}$ in the meta-training stage, we aim to learn the knowledge which can be adapted to novel tasks in the meta-test stage. Each task $\mathcal{T}$ is drawn from a few-shot labeled dataset $\mathcal{D}$. The $\mathcal{U}$ and $\mathcal{D}$ are drawn from the same distribution but a different set of classes. Specially, a $K$-way $S$-shot classification task $\mathcal{T}$  consists of support data $\mathcal{S}=\{(\mathbf{x}_s, \mathbf{y}_s)\}_{s=0}^{KS}$ with $K$ classes of $S$ few labeled samples and query data $\mathcal{Q}=\{\mathbf{x}_q\}_{q=0}^{Q}$ with $Q$ unlabeled samples. Our goal is to  predict the labels of $\mathcal{Q}$ given $\mathcal{S}$.

\subsection{Causal Insight} \label{sec:causal insight}

Unsupervised meta-learning methods are confused by the context-bias. To analyze how the bias arises, we formulate the problem into SCM in Figure \ref{fig:intro_scm}.
In the SCM, 1) $D \to X$ means that the priors D determine where the object appears in an image, \eg, the context-priors in training images of Figure \ref{fig:intro} put the bird object in the sky. 2) $D \to Y$ denotes that the priors D affect the predictions $Y$, \eg, the wing and sky priors lead to the bird prediction. 3) $D_1,\cdots,D_d$ are dependent statistically, \eg,  the ``sky'', ``wing'' and  prior are not independent but causally dependent. Their causal relationships need to be determined (dashed lines).  4) $X \to Y$ is the regular classification process. 

From the SCM, we observe that context-priors $D$ confound the effect  that input $X$ has on prediction $Y$, which leads to the bias. Thus, it is critical to eliminate the confounding effects, we then apply causal intervention with the do-operator \cite{glymour2016causal} as follows (Details in Supp. 3.1), 
\begin{align}\small
	P(\mathbf{y}|do(\mathbf{x})) = & \sum_{{\dbar}_1,  \cdots, {\dbar}_d}P(\mathbf{y}|\mathbf{x}, D_1={\dbar}_1, \cdots, \notag \\  
	&D_d={\dbar}_d)P(D_1={\dbar}_1,  \cdots, D_d={\dbar}_d) \label{eq:do}
\end{align}
where ${\dbar}_i$ ranges over all values that variables $D_i$ can take. 

Equation \ref{eq:do} informs that intervening on $\mathbf{x}$ calls for the joint distribution of $D$. Note that $D_1, \cdots, D_d$ may be dependent statistically (\ie, $P(D) \neq \Pi_{i=1}^d P(D_i)$). Inspired by CCP \cite{scholkopfCausalRepresentationLearning2021}, we assume the common causes are ones of priors. Then finding the common causes suggests discovering the causal relationships among the priors. The causal relationships can be represented by a DAG (dashed lines). For example in Figure \ref{fig:intro}, the flying prior is the common cause of sky and wing priors, the DAG is ``sky $\leftarrow$ flying $\rightarrow$ wing", and the latter two are independent when conditioned on the flying. Furthermore, based on ICM \cite{scholkopfCausalRepresentationLearning2021}, the joint distribution $P(D)$ can be factorized into, 
\begin{equation}
	P(D)=\prod_{i=1}^d P(D_i|\operatorname{PA}(i)) \label{eq:factorization}
\end{equation}
where $\operatorname{PA}(i)$ denotes the parents of $D_i$, which can be obtained from the DAG. 

\begin{figure}[t]
	\centering
	\includegraphics[width=0.7\linewidth]{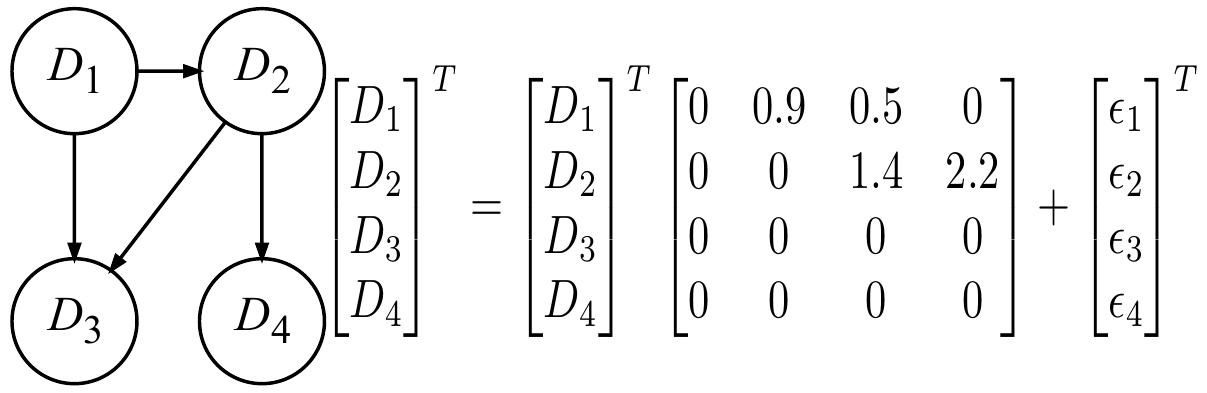}
	\caption{An SEM. [Left]:  DAG with 4 nodes. [Right]: A linear equation for Gaussian SEM with noise $\epsilon \sim \mathcal{N}(0, \bm{I})$. }
	\label{fig:sem}
\end{figure}	

To discover the DAG, we utilize Gaussian Structural Equation Model (SEM) \cite{pearl2000models}. Figure \ref{fig:sem} shows a linear-Gaussian SEM. Formally, given the variables $D$, there exist functions $h_i$ and $h_j:\mathbb{R}^{d} \rightarrow \mathbb{R}$ such that
\begin{equation}
	\label{equ:sem}
	D_i = h_i(D) + U_i, \quad D_j = h_j(D) + U_j
\end{equation}
where $U_i$ and $U_j$ are independent Gaussian noises, and $h_i$ and $h_j$ are regarded as structural functions.
The relationship between $h_i$ and $\operatorname{PA}()$ is that $h_i(\dbar_1, \cdots, \dbar_d)$ does not depend on $\dbar_k$ if $D_k \notin \operatorname{PA}(i)$. 

The DAG can be learned by maximum likelihood estimation $\mathbb{E}[D_i|h_i(D)]$ and $\mathbb{E}[D_j|h_j(D)]$ over $D$. Its ``DAGness'' can be enforced using a trace exponential regularizer such as NoTears penalization \cite{Zheng2018dags}.
Insufficient penalization weight may not ensure the ``DAGness'' and weaken the effect of bias-removal, but default weight works for most scenarios. If the causal graph is Non-DAG graph, a solution is to learn such mixed graphs with score-based methods \cite{pmlr-v108-bernstein20a}. It is compatible with our method.

\subsection{Adjustment Formulation} 
This section offer an adjustment formulation for Equation \ref{eq:do}.
Specially, given the DAG function $h=\{h_i\}_{i=1}^{d}$, the distribution $P(D)$ is approximated by $P(D_1=\dbar_1,\cdots, D_d=\dbar_d) \approx P(D=\bm{\dbar}|h(D=\bm{\dbar}))$, where $\bm{\dbar}=[ \dbar_1|\cdots| \dbar_d ]\in \mathbb{R}^{1 \times d}$. 
Also, given the input data $\mathbf{x}$, we assume its latent codes $\mathbf{z}\in \mathbb{R}^{1\times d} $ via VAE \cite{Kingma2014vae}. Since VAE has been shown to achieve some useful disentangling \cite{higgins2016beta}, we perform the dimensional-wise product to make each latent code represents one prior, \ie, $\mathbf{z} \leftarrow \mathbf{z} \otimes \mathbf{\dbar}$. Then we have $P(D=\bm{\dbar}|h(D=\bm{\dbar}))=P(Z=\mathbf{z}|h(Z=\mathbf{z}))$. Finally, the adjustment formulation yields,
\begin{equation} 
	p(\mathbf{y}|do(\mathbf{x}))=  \mathbb{E}_{\underbrace{p(\mathbf{z}|\mathbf{x})}_{\text{Sampling}}}\mathbb{E}_{\underbrace{p(\mathbf{z}|h(\mathbf{z}))}_{\text{Adjusting}}} p(\mathbf{y}|\mathbf{z}) \label{eq:backdoor}
\end{equation}
Equation \ref{eq:backdoor} reveals that the causal intervention can be accomplished by the sampling term  $p(\mathbf{z}|\mathbf{x})$  and the adjusting term $p(\mathbf{z}|h(\mathbf{z}))$ with the DAG function $h$. Note that the adjusting term is short for two steps: 1) Draw $\mathbf{e} \sim p(\mathbf{e} | \mathbf{z})$; 2) Make $\mathbf{z}-\mathbf{e} \sim \mathcal{N}(0, \bm{I})$, which is a constraint that forces $h$ to follow the DAG in $\mathbf{z}$. Thus, we call it adjusting.

While variables $\mathbf{z}$ and function $h$ may be non-identifiable due to non-conditional additionally observed variables (\eg, DAG label) \cite{khemakhem2020variational}, we can choose suitable inductive biases to recover a certain structure in the real world \cite{locatello2019challenging, pmlr-v139-trauble21a}. Besides, the formulation is also sufficient for classification based on the two causal principles. Empirical results in Section \ref{sec:count} also reveal some meaningful explanation. 

Though \cite{yang2021causalvae, Kim_Shin_Jang_Song_Joo_Kang_Moon_2021} have studied learning causality with VAE, their generative process is ``noises $\rightarrow$ causal codes $\rightarrow$ images''and needs additional observations to learn the distributution of codes, which is impractical and limited. While our generative is ``noise $\rightarrow$ images'' and make ``noise $=$ causal codes''. It is as flexible as vanilla  VAE. Compared to Deconfounder \cite{wang2019blessings}, our causal structure on the latent confounders is defined and to be learned by DAG learning methods.

\subsection{Causal Latent Space}  \label{sec:method-Causal} 
To achieve the downstream task such as {clustering} and {classification}, we introduce the causal latent space (CaLS) and study the computation of weighted sum in this space. Particularly, we assume the distribution of the causally independent codes $\mathbf{z} \in \mathbb{R}^{1\times d}$ is Gaussian \footnote{Actually we assume the error term $\bm{\epsilon}=\mathbf{z}-h(\mathbf{z})$ is Gaussian and ignore this error to focus on $\mathbf{z}$ and the corresponding space.}
\begin{equation}
	\mathbf{z} \sim \mathcal{N}( h(\mathbf{z}), \bm{I}) \label{equ:gau}
\end{equation}	    
We refer to the latent codes in CaLS as causal codes, and the causal codes follow the same DAG. 
Then, the weighted sum latent codes can be obtained by the following proposition.
\begin{proposition} 
	\label{sec:prop1}
	Assume there are $n$ causal codes $\mathbf{Z}\in \mathbb{R}^{n \times d}$ shared same $h$ that represents the DAG, an assignment $\bm{w} \in \mathbb{R}^{n\times1}$ satisfying $\bm{w}^T\bm{1}=1$ and the weighted sum $\overline{\mathbf{z}}=\bm{w}^{T}\mathbf{Z}$. Then  
	\begin{equation}
		\overline{\mathbf{z}} \sim \mathcal{N}( h(\overline{\mathbf{z}}),  \bm{w}^T\bm{w} \bm{I} )
		\label{eq:prop1}
	\end{equation}
	whenever $h$ is linear or non-linear function. 
	Proof is available in Supp. 3.2. 	  
\end{proposition} 
Proposition \ref{sec:prop1} shows that whatever the function $h$, the causal relationships of $\overline{\mathbf{z}}$ by weighted sum over $\mathbf{z}$ will remain unchanged  as $h$ can express the DAG structure.

\section{Causal Meta VAE}

\begin{figure}[t]
	\centering
	\begin{subfigure}{.27\textwidth}
		\centering
		\includegraphics[width=\linewidth]{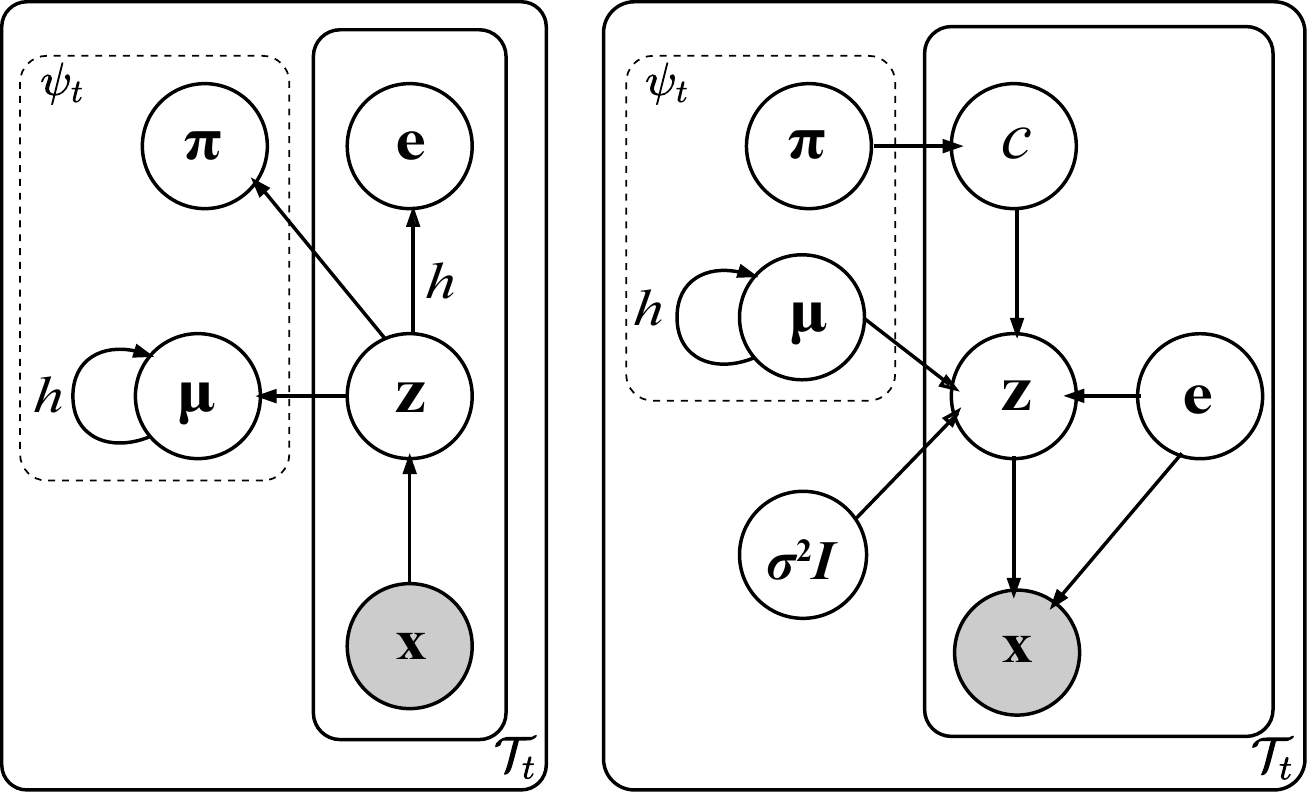}
		\caption{}
		\label{fig:intro_a}
	\end{subfigure}
	\hfill
	\begin{subfigure}{.147\textwidth}
		\centering
		\includegraphics[width=\linewidth]{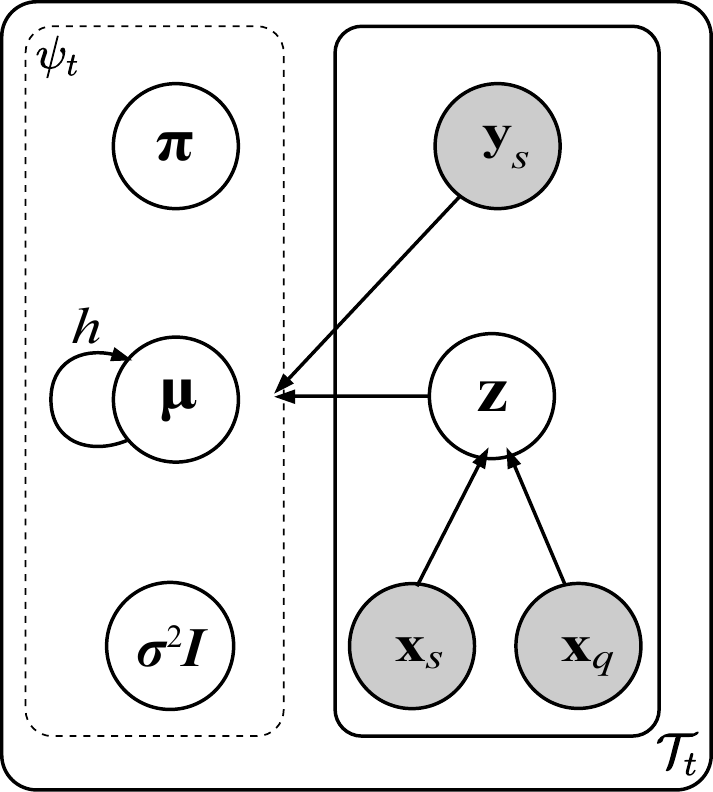}
		\caption{}
		\label{fig:intro_b}
	\end{subfigure}	
	\caption{Graphical model of CMVAE. (a) Unsupervised meta-training. CMoG prior $\psi_t=\{\bm{\pi}, \bm{\mu}\}$. [Left] Variational posterior $q_\phi(\mathbf{z}|\mathbf{x}, \mathcal{T}_t)$, $q_\phi(\mathbf{e}|\mathbf{z},\mathbf{x})$. $\psi_t$ is learned by causal-EM. [Right] Generative
		model $p_\theta(\mathbf{x}|\mathbf{z},\mathbf{e})$, $p(\mathbf{z}|\mathbf{e})$. 
		(b)Meta-test by semi-supervised causal-EM. } 
	\label{fig:cmvae}
	
\end{figure}

To demonstrate the effectiveness of pipeline, we extent the baseline \cite{lee2021metagmvae} into our CMVAE. It includes the Causal Mixture of Gaussian (CMoG), unsupervised meta-training and meta-test methods with novel causal Expectation Maximization. The following notation subscript is used: $\mathbf{z}_{[i]} \in \mathbb{R}^{1 \times d}$ for $i$-th observation of $\mathbf{Z}$, and $\mathbf{z}_j \in \mathbb{R}^{n \times 1}$ for $j$-th dimension of $\mathbf{Z}$. Figure \ref{fig:cmvae} shows the graphical model of CMVAE.

\subsection{Causal Mixture of Gaussians} 
The Causal Mixture of Gaussians (CMoG) is an extension of MoG distribution in the CaLS  based on proposition \ref{sec:prop1},
\begin{gather}
	c \sim \operatorname{Cat}(\bm{\pi}), \quad \mathbf{z}|c  \sim \mathcal{N}(\bm{\mu}_{[k]}, {\bm{\sigma}}_{[k]}^2\bm{I}) ,   \notag \\  	\bm{\mu}_{[k]} \sim \mathcal{N}(h(\bm{\mu}_{[k]}), s_k^2\bm{I})	\label{eq:reg}
\end{gather}
where $\bm{\pi}$ is $K$ dimensional weights, $(\bm{\mu}_{[k]}, \bm{\sigma}_{[k]}^2)$ are mean and diagonal covariance of the $k$-th mixture modality, and the scalar $s_k^2$  is a scaling parameter. Here we take the diagonal covariance $\bm{\sigma}_{[k]}^2\bm{I}$ instead of $\bm{\Sigma}_k$ since the relationships between dimensions can be mined by learning the DAG. From another perspective, Eq. \ref{eq:reg} can be seen as a regularization to make the modality causally independent. we refer to it as {causal modality}.

\subsection{Unsupervised Meta-training} 

We now describe unsupervised meta-training in causal latent space based on VAE \cite{Kingma2014vae}.  Given a meta-training task $\mathcal{T}_t=\{\mathbf{x}^i \in \mathcal{U} \}_{i=1}^M$, the goals are to optimize the variational lower bound of the data marginal likelihood of task $\mathcal{T}_t$ using an variational posterior. 
Specifically, for the unsupervised meta-learning where labels are unknown, we define the variational posterior $q_\phi(\mathbf{z}|\mathbf{x}, \mathcal{T}_t)$ and the task-specific CMoG priors $p_{\psi_t^{*}}(\mathbf{z})$.  For learning the causal structure, let $\mathbf{e}$ be sampled from the causal latent space, where function $h$ is applied to $\mathbf{z}$, \ie,   $\mathbf{e}|\mathbf{z} \sim \mathcal{N}(h(\mathbf{z}), \bm{I})$. For posterior network, we use a factorization $q_\phi(\mathbf{e},\mathbf{z}|\mathbf{x},\mathcal{T}_t)=q_\phi(\mathbf{e}|\mathbf{z},\mathbf{x},\mathcal{T}_t)q_\phi(\mathbf{z}|\mathbf{x},\mathcal{T}_t)$, sampling $\mathbf{z}$ given $\mathbf{x} \in \mathcal{T}_t$ first, then conditionally sampling $\mathbf{e}$ based on these values. It leads to the evidence lower bound (ELBO) (Details in Supp. 3.3),
\begin{align} 
	\label{eq:elbo}
	\mathbb{E}_{q_\phi(\mathbf{z}|\mathbf{x}, \mathcal{T}_t)}[
	\mathbb{E}_{q_\phi(\mathbf{e}|\mathbf{z},\mathbf{x})}[ 
	\log p_\theta(\mathbf{x}|\mathbf{z},\mathbf{e}) - \log \frac{q_\phi(\mathbf{e}|\mathbf{z},\mathbf{x})}{p(\mathbf{e}|\mathbf{z})} ]	\notag \\
	+\log  p_{\psi_t^{*}}(\mathbf{z})-\log q_\phi(\mathbf{z}|\mathbf{x}, \mathcal{T}_t)]
\end{align}
where $x\in \mathcal{T}_t$. The ELBO can be approximated by Monte Carlo estimation. We then describe these variational posteriors and priors in detail.

\noindent\textbf{Variational Posterior.}
The task-conditioned variational posterior $q_\phi(\mathbf{z}|\mathbf{x}, \mathcal{T}_t)$ is to encode the dependency into the latent space between data in current task. Following \cite{lee2021metagmvae}, we take task $\mathcal{T}_t$ as inputs and denote,
\begin{gather}
	H=\operatorname{TE}(F(\mathbf{x})),  \mathbf{x} \in \mathcal{T}_t, \quad
	\bm{\mu} = W_{\bm{\mu}}H + b_{\bm{\mu}}, \quad  \notag \\ \bm{\sigma}^2=\exp(W_{\bm{\sigma}^2}H+b_{\bm{\sigma}^2}),
	q_\phi(\mathbf{z}|\mathbf{x}, \mathcal{T}_t) = \mathcal{N}(\mathbf{z}|\bm{\mu},\bm{\sigma}^2) \label{eq:q_phi}
\end{gather} 
where $\operatorname{TE}(\cdot)$ is multi-head self-attention mechanism \cite{vaswani2017attention}, $F$ is a convolutional neural network (or an identity function). To learn the causal structure, we apply the function $h$ to the latent space and then sample $\mathbf{e}$ from the obtained causal latent space, 
\begin{equation}
	q_\phi(\mathbf{e}|\mathbf{z},\mathbf{x}) = \mathcal{N}(\mathbf{e}| h(\mathbf{z}), \bm{I}), \quad \mathbf{z} \sim q_\phi(\mathbf{z}|\mathbf{x}, \mathcal{T}_t) \label{eq:q_h}
\end{equation}

\noindent\textbf{Causally Conditional Prior.}
Ideally if the DAG $h$ represents the true causal structure, the conditional prior $p(\mathbf{e} | \mathbf{z})$ can be obtained by replacing the unknown $h$,
\begin{equation}
	\label{eq:h_prior}
	p(\mathbf{e} | \mathbf{z})  = \mathcal{N}(0, \bm{I})+ h(\mathbf{z}) 
	= \mathcal{N}(\mathbf{e} | \mathbf{z}, 2\bm{I})
\end{equation}	 

\begin{algorithm}[t]
	\small
	\caption{Unsupervised Causal Meta-training}
	\label{alg:meta-training}
	\begin{algorithmic}
		\STATE {\bfseries Input:} An unlabeled dataset $\mathcal{U}$, causal-EM steps \texttt{step}.
		\STATE Initialized parameterized $q_\phi$, $p_\theta$.
		\WHILE {\textit{not converged}} 
		\STATE Generate unlabeled task $\mathcal{T}_t=\{\mathbf{x}_u | \mathbf{x}_u \in \mathcal{U}\}$
		\STATE Draw $\mathbf{z} \sim q_\phi(\mathbf{z}|\mathbf{x}, \mathcal{T}_t)$, $\mathbf{e} \sim q_\phi(\mathbf{e}|\mathbf{z},\mathbf{x}) $ in Eq. \ref{eq:q_phi}, \ref{eq:q_h}
		\STATE Compute $\psi_t^*$ in Eq. \ref{eq:em-train} with \texttt{step} causal-EM 
		\STATE Compute loss $\mathcal{L}$ in Eq. \ref{eq:loss} and update $\phi$, $\theta$, $h$				
		\ENDWHILE
	\end{algorithmic}
\end{algorithm}

\noindent\textbf{Task-specific Prior.} 
The task-specific causal multi-modal prior is modeled via CMoG and formally factorized as:
\begin{gather} 			
	p_{\psi_t}(\mathbf{z})=\sum_{c=0}^{K}p_{\psi_t}(\mathbf{z}|c)p_{\psi_t}(c), \quad p_{\psi_t}(c)=\operatorname{Cat}({c|\bm{\pi} }), \notag \\ 
	p_{\psi_t}(\mathbf{z}|c) =  \mathcal{N}(\mathbf{z}|\bm{\mu}_{[k]}, {\bm{\sigma}}_{[k]}^2\bm{I}) \mathcal{N}(\bm{\mu}_{[k]}|h(\bm{\mu}_{[k]}), s_k^2\bm{I})
	\label{eq:task-param}		
\end{gather} 		
where the task-specific parameters $\psi_t$ is defined as $\psi_t =\{\bm{\pi}, \bm{\mu}_{[k]}, {\bm{\sigma}}_{[k]}^2\bm{I}, s_k^2 \}$. Maximizing ELBO in Eq. \ref{eq:elbo} results in locally maximizing the following maximum causal posterior (MCP) problem:
\begin{equation}
	\psi_t^{*} = \operatornamewithlimits{argmax}_{\psi_t}\sum\log p(\psi_t|\mathbf{z}) 
\end{equation}	
Without losing the DAG structure, the derived EM equations in closed forms are referred to as \textit{causal-EM} (Derivations in Supp. 3.4),
\begin{gather}
	\textbf{E: }  \omega_{ik} = \frac{\alpha_k \mathcal{N}(\mathbf{z}_{[i]} | \bm{\mu}_{[k]}, \bm{I}) \mathcal{N}(\bm{\mu}_{[k]} | h(\bm{\mu}_{[k]}), \gamma^2\bm{I}) }{\sum_k\alpha_k \mathcal{N}(\mathbf{z}_{[i]} | \bm{\mu}_{[k]}, \bm{I}) \mathcal{N}(\bm{\mu}_{[k]} | h(\bm{\mu}_{[k]}), \gamma^2\bm{I}) } \quad \notag \\
	\textbf{M: } 	\bm{\mu}_{[k]} = \frac{\sum_{i=1}^M\omega_{ik}\mathbf{z}_{[i]}(\bm{I} + \epsilon(\gamma^{-1}\bm{I})\epsilon^T(\gamma^{-1}\bm{I}) )^{-1}}{\sum_{i=1}^M\omega_{ik}}  \label{eq:em-train} 
\end{gather}
where $\epsilon(\mathbf{z}) = \mathbf{z} -h(\mathbf{z}) $ and $\alpha_k = \frac{\sum_{i=1}^M\omega_{ik}}{\sum_{k=1}^K\sum_{i=1}^M\omega_{ik}}$. It can also be simplified using the inverse covariance matrix and we want to show that the term $\epsilon(\gamma^{-1}\bm{I})$ allows that $j'$ propagates its information to ${j}$ if ${j'} \in \operatorname{PA}(j)$, then intervenes and refines $\bm{\mu}_{[k]}$.  Following the assumption of VAE, the covariance of Gaussian distribution is set to $\bm{I}$. We also observe that setting $s_k^2$ to a fixed hyper-parameter $\gamma^2$ results in better convergence. The $\alpha_k$ is initialized as $\frac{1}{K}$, and $\bm{\mu}_{[k]}$ is initialized as:
$ 		 \bm{\mu}_{[k]} = \frac{\sum_i^K\mathbf{z}_{[i]}
	(\bm{I} + \epsilon(\gamma^{-1}\bm{I})\epsilon^T(\gamma^{-1}\bm{I}) )^{-1}}{K} \label{eq:init_u_train} 	$
where $\{\mathbf{z}_{[i]}\}_{i=1}^K$ are randomly selected. By performing a few causal-EM steps iteratively, the MCP converges and task-specific parameters $\psi_t^*$ is obtained. 

\subsection{Training Objective}

\begin{figure}
	\centering
	\begin{subfigure}[t]{.11\textwidth}
		\centering
		\includegraphics[width=\linewidth]{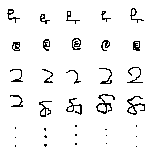}
		\caption{}
		\label{fig:res_a}
	\end{subfigure}
	\hfill
	\begin{subfigure}[t]{.11\textwidth}
		\centering
		\includegraphics[width=\linewidth]{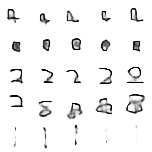}
		\caption{}
		\label{fig:res_b}
	\end{subfigure}		
	\hfill		
	\begin{subfigure}[t]{.11\textwidth}
		\centering
		\includegraphics[width=\linewidth]{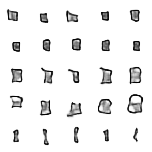}
		\caption{}
		\label{fig:res_c}
	\end{subfigure}
	\hfill
	\begin{subfigure}[t]{.11\textwidth}
		\centering
		\includegraphics[width=\linewidth]{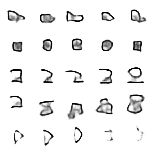}
		\caption{ }
		\label{fig:res_d}
	\end{subfigure}
	\caption{Visualization on Omniglot. (a, b) The samples and generated samples for each mode at supervised meta-test step of CMVAE. Each row stands for each modality obtained by EM. (c, d) Counterfactual samples by intervention on {causes} and {effects}, respectively. The larger the change, the better the intervention, the more we can show that our method has learned the causes and effects.}  
\end{figure}

\textbf{DAG Loss.} DAG loss is to ensure the ``DAGness''. We consider two types. 1) Linear SEM, $h(\mathbf{z})= \mathbf{z}\mathbf{A}$, where $\mathbf{A}\in \mathbb{R}^{d\times d}$. 2) Nonlinear SEM, we model it with a multilayer perceptron (MLP), $ h_i(\mathbf{z})=\sigma(\sigma(\sigma(\mathbf{z}\mathbf{W}_i^1) \cdots) \mathbf{W}_i^l)$, and define $[\mathbf{A}]_{mi}=\|m\operatorname{th-row}(\mathbf{W}_i^1) \|_2$ where  $\|\cdot\|_2$ is $\ell _2$ norm. Then the DAG loss \cite{Zheng2018dags} is 	
\begin{equation}
	\mathcal{R}_{D}(\mathbf{A})= (\operatorname{tr}(\exp(\mathbf{A} \circ \mathbf{A}))-d)^2
\end{equation} 

\noindent\textbf{Objective.} After getting the task-specific parameters $\psi_t^*$, we use gradient descent-based method \textit{w.r.t.} the variational parameter $\phi$, the generative parameter $\theta$ and the parameters of function $h$ and minimize the following objective,
\begin{equation}
	\mathcal{L} = -\text{ELBO} + \lambda_1 \mathcal{R}_{D}(\mathbf{A}) + \lambda_2 \|\mathbf{A}\|_1 \label{eq:loss}
\end{equation}
where $\lambda_1, \lambda_2$ are hyper parameters which control the ``DAGness", and $\|\cdot\|_1$ is $\ell _1$ norm. 

Algorithm \ref{alg:meta-training} shows the steps of the unsupervised meta-training stage. The outputs of unsupervised meta-training stage consists of variational
parameter $\phi$, the generative parameter $\theta$ and the parameters of function $h$. Similar to the regular meta-training stage, these outputs are also model initialization as it is a bi-level optimization \cite{9638340, pmlr-v162-vicol22a}.  The inner optimization is to maximize ELBO over task-specific $\psi$ in Equation \ref{eq:elbo}. In th outer loop, our method is to minimize the loss with regard to  task-agnostic parameters $\phi$, $\theta$ and $h$ in Equation \ref{eq:loss}.

\begin{table*}[t]
	\centering		
	\caption{Results (way, shot) in Omniglot and \textit{mini}ImageNet. The ACAI/DC (RO/N) mean ACAI clustering (Random Out-of-class samples) on Omniglot and DeepCluster (Noise) on \textit{mini}ImageNet. }
	\small
	\scalebox{0.9}{
		\begin{tabular}{llcccc|cccc}
			\toprule
			&            &             \multicolumn{4}{c}{Omniglot (way, shot)}              &       \multicolumn{4}{c}{\textit{mini}ImageNet (way, shot)}        \\ \toprule
			Method                           & Clustering &     (5,1)      &     (5,5)      &     (20,1)     &     (20,5)     &     (5,1)      &     (5,5)      &     (5,20)     &     (5,50)     \\ \midrule
			\textit{Training from Scratch}   & N/A        &     52.50      &     74.78      &     24.91      &     47.62      &     27.59      &     38.48      &     51.53      &     59.63      \\ \midrule
			CACTUs-MAML                      & BiGAN      &     58.18      &     78.66      &     35.56      &     58.62      &     36.24      &     51.28      &     61.33      &     66.91      \\
			CACTUs-ProtoNets                 & BiGAN      &     54.74      &     71.69      &     33.40      &     50.62      &     36.62      &     50.16      &     59.56      &     63.27      \\
			CACTUs-MAML                      & ACAI/DC    &     68.84      &     87.78      &     48.09      &     73.36      &     39.90      &     53.97      &     63.84      &     69.64      \\
			CACTUs-ProtoNets                 & ACAI/DC    &     68.12      &     83.58      &     47.75      &     66.27      &     39.18      &     53.36      &     61.54      &     63.55      \\
			UMTRA                            & N/A        &     83.80      &     95.43      &     74.25      & \textbf{92.12} &     39.93      &     50.73      &     61.11      &     67.15      \\
			LASIUM-MAML-RO/N                 & N/A        &     83.26      &     95.29      &       -        &       -        &     40.19      &     54.56      &     65.17      &     69.13      \\
			LASIUMs-ProtoNets-RO/N           & N/A        &     80.12      &     91.10      &       -        &       -        &     40.05      &     52.53      &     59.45      &     61.43      \\
			Meta-GMVAE                       & N/A        &     94.92      &     97.09      &     82.21      &     90.61      &     42.82      &     55.73      &     63.14      &     68.26      \\
			IFSL$^\dagger$                   & N/A        &     94.22      &     97.01      &     82.21      &     90.65      &     42.90      &     56.01      &     63.24      &     68.90      \\ \midrule
			CMVAE (\textit{ours})            & N/A        & \textbf{95.11} & \textbf{97.14} & \textbf{82.58} &    {90.79}     & \textbf{44.27} & \textbf{58.95} & \textbf{66.25} & \textbf{70.54} \\ \midrule
			MAML (\textit{Supervised})       & N/A        &     94.46      &     98.83      &     84.60      &     96.29      &     46.81      &     62.13      &     71.03      &     75.54      \\
			ProtoNets  (\textit{Supervised}) & N/A        &     98.35      &     99.58      &     95.31      &     98.81      &     46.56      &     62.29      &     70.05      &     72.04      \\ \bottomrule
		\end{tabular}
	}
	\label{tab:rest_o_m}
\end{table*}

\subsection{Supervised Meta-test}

With CMoG priors, each causal modality can be seen as a pseudo-class concept. To adapt the causal modality to few-shot classification, we use both support set and query set and draw causal latent codes from the variational posterior $q_\phi$. During the meta-test given a task $\mathcal{T}=\{(\mathcal{S},\mathcal{Q}) | \mathcal{S}=\{\mathbf{x}_s, \mathbf{y}_s\}_{s=1}^S, \mathcal{Q}=\{\mathbf{x}_q\}_{q=1}^Q \}$, the goal is to compute the conditional probability $p(\mathbf{y}_q|\mathbf{x}_q, \mathcal{T})$ \textit{w.r.t.}  variational posterior $q_\phi$, the causal multi-modal prior parameter $\psi^{*}$ and the backdoor adjustment in Equation \ref{eq:backdoor}:
\begin{equation}
	\label{eq:meta-test}
	p(\mathbf{y}_q|\mathbf{x}_q, \mathcal{T}) = \mathbb{E}_{q_\phi(\mathbf{z}_q|\mathbf{x}_q, \mathcal{T})p(\mathbf{z}_q|h(\mathbf{z}_q))}[p_{\psi^*}(\mathbf{y}_q|\mathbf{z}_q)]
\end{equation}  
Eq. \ref{eq:meta-test} can also be computed by Bayes rule and Monte Carlo sampling. Then the predicted label is
\begin{equation} \label{eq:pred_y}
	\hat{\mathbf{y}}_q=\operatornamewithlimits{argmax}_kp(\mathbf{y}_q=k|\mathbf{z}_q,\mathcal{T})
\end{equation}

To obtain the optimal prior parameters $\psi^{*}$ in current meta-test task $\mathcal{T}$ and make the causal modality as label, we develop a semi-supervised causal-EM algorithm. In particular, we sample the causal  code $\mathbf{z} \sim q_\phi(\mathbf{z}|\mathbf{x},\mathcal{T})$ first and then get the causal multi-modalities with steps as follows,
\begin{gather} 
	\textbf{E: } \omega_{qk} = \frac{ \mathcal{N}(\mathbf{z}_{[q]} | \bm{\mu}_{[k]}, \bm{\sigma}_{[k]}^2) \mathcal{N}(\bm{\mu}_{[k]} | h(\bm{\mu}_{[k]}), \gamma^2\bm{I}) }{\sum_k \mathcal{N}(\mathbf{z}_{[q]} | \bm{\mu}_{[k]},  \bm{\sigma}_{[k]}^2) \mathcal{N}(\bm{\mu}_{[k]} | h(\bm{\mu}_{[k]}), \gamma^2\bm{I}) } \notag \\	  		  		
	\textbf{M: } \text{ }  \quad 
	\tilde{\bm{\mu}}_{[k]} = 
	\sum_s\tilde{\omega}_{sk}\mathbf{z}_{[s]} +  \sum_q\tilde{\omega}_{qk}\mathbf{z}_{[q]} \quad \quad \quad \quad \quad \quad \notag \\
	{\bm{\mu}}_{[k]} = \tilde{\bm{\mu}}_{[k]} (\bm{I} + \epsilon(\gamma^{-1}\bm{\sigma}_{[k]})\epsilon^T(\gamma^{-1}\bm{\sigma}_{[k]}) )^{-1} \label{eq:em-test}   \\
	\bm{\sigma}_{[k]}^2 = \sum_s\tilde{\omega}_{sk}(\mathbf{z}_{[s]} - \bm{\mu}_{[k]})^2 +   \sum_q\tilde{\omega}_{qk}(\mathbf{z}_{[q]} - \bm{\mu}_{[k]})^2  \notag
\end{gather}
where  $	\tilde{\omega}_{sk} =\frac{\mathds{1}_{\mathbf{y}_s=k}}{\sum_s\mathds{1}_{\mathbf{y}_s=k}+\sum_q\omega_{qk}}, 	  	 \tilde{\omega}_{qk}= \frac{\omega_{qk}}{\sum_s\mathds{1}_{\mathbf{y}_s=k}+\sum_q\omega_{qk}}$ and $\mathds{1}$ is the indicator function. We keep the mixture probability fixed to $\frac{1}{K}$ due to the uniformly distributed labels and use diagonal covariance $\bm{\sigma}_{[k]}^2$ instead of $\bm{I}$ to obtain more accurate results. The $\bm{\mu}_{[k]}$ is initialized as:
$
\bm{\mu}_{[k]} = \frac{\sum_s\mathds{1}_{\mathbf{y}_s=k}\mathbf{z}_{[s]}
	(\bm{I} + \epsilon(\gamma^{-1}\bm{I})\epsilon^T(\gamma^{-1}\bm{I}) )^{-1}}{\sum_s\mathds{1}_{\mathbf{y}_s=k}} \label{eq:init_u_test}
$
Finally, we can get the solution of MCP and $\psi^*$  by a few steps iteratively similar to the meta-training. 

\section{Experiment} 
In this section we show the empirical performance of our method on few-shot classification tasks.

\subsection{Experiment Settings}

\begin{figure}[]   
	\centering
	\begin{subfigure}[t]{.15\textwidth}
		\centering
		\includegraphics[width=\linewidth]{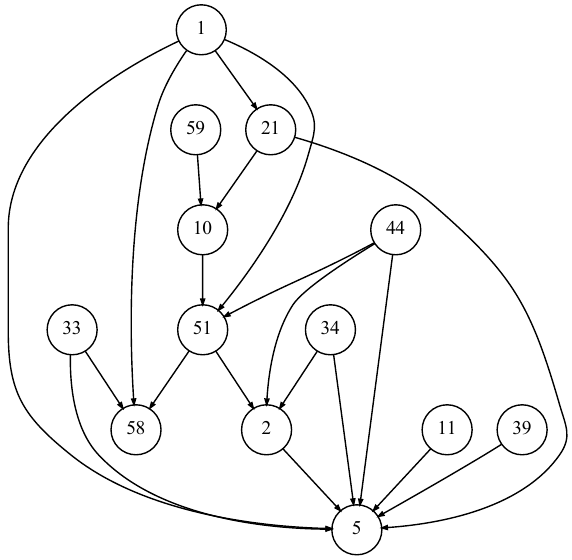}
		\caption{}
		\label{fig:dag_a_full}
	\end{subfigure}
	\hfill
	\begin{subfigure}[t]{.15\textwidth}
		\centering
		\includegraphics[width=\linewidth]{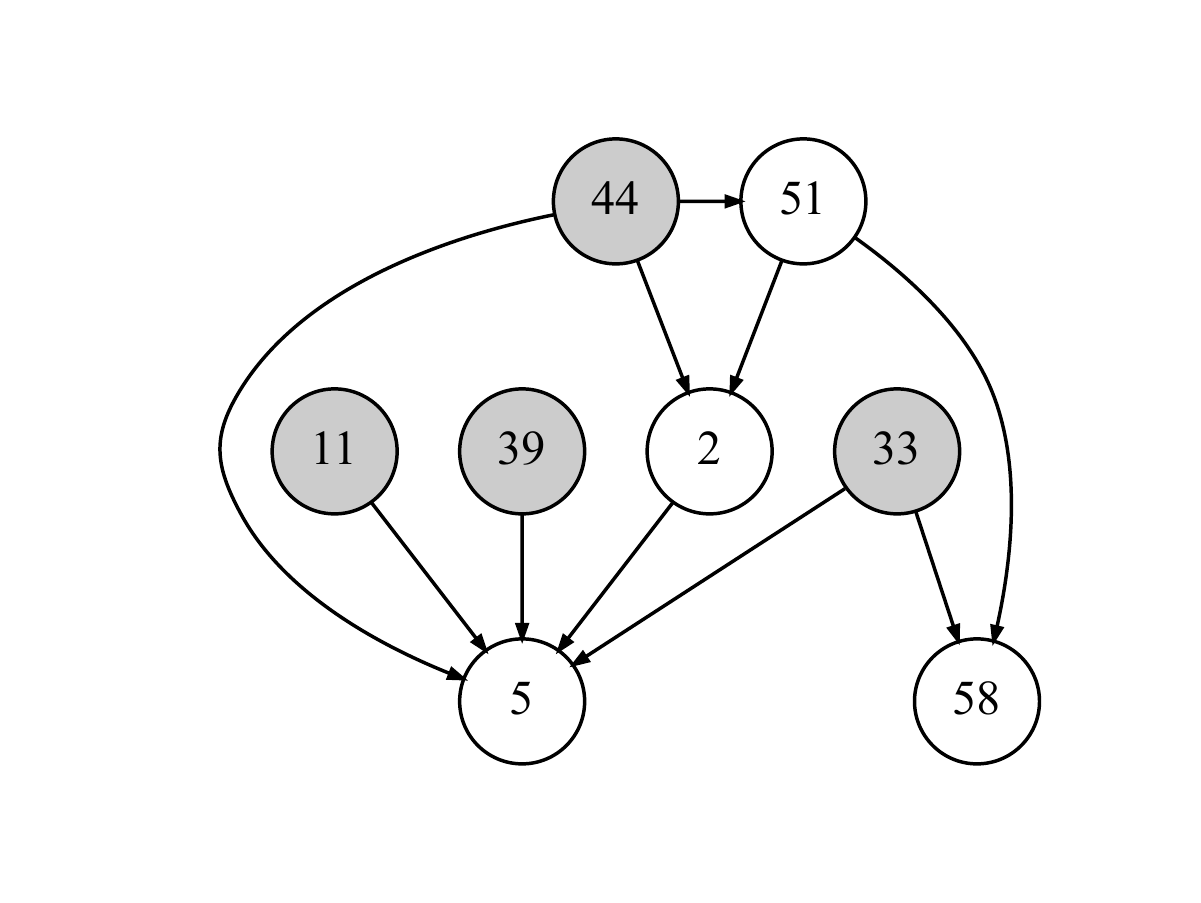}
		\caption{}
		\label{fig:dag_a}
	\end{subfigure}
	\hfill
	\begin{subfigure}[t]{.15\textwidth}
		\centering
		\includegraphics[width=\linewidth]{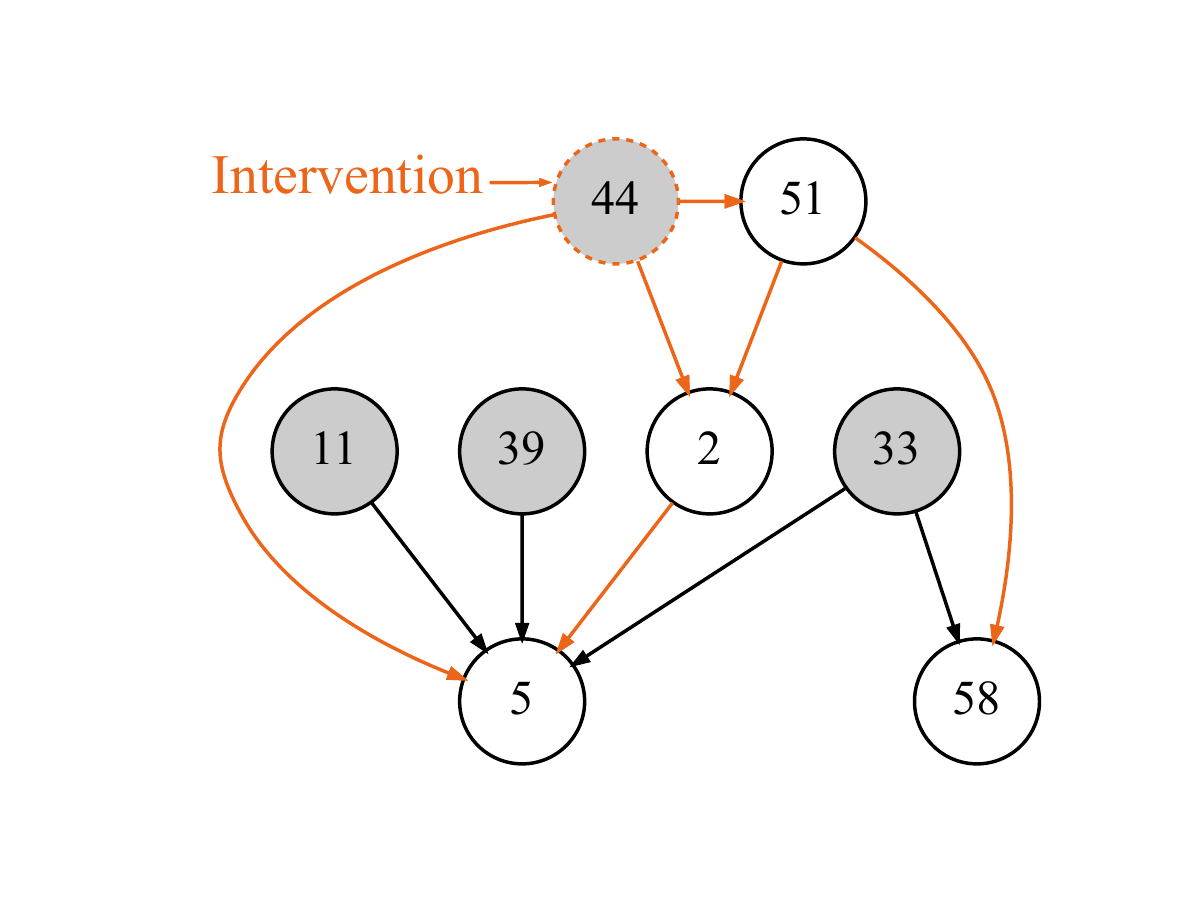}
		\caption{}
		\label{fig:dag_b}
	\end{subfigure}	
	\caption{ (a) DAG on Omniglot by the learned $\mathbf{A}$. Each node represents each dimension of $\mathbf{z}$. Other nodes are not shown because they are independent and have no cause-to-effect relationship (b) Part of DAG to show the causes and effects. The gray nodes represent the causes. (c) Intervention on one cause, \eg, $\mathbf{z}_{44}$, will change the effects \eg, $\mathbf{z}_{5}$ while will not change other causes, \eg, $\mathbf{z}_{11}$. Best viewed in color.}	
	\label{fig:dags}
	
\end{figure}

\noindent\textbf{Dataset.} One biased toy dataset and three natural datasets are used to test our algorithm. \textbf{1) Toy dataset.} It is a 2-way biased dataset with a synthetic "bird" and "plane" image. (Details in Supp. 5.1.)  \textbf{2) Omniglot.} Omniglot consists of 1,623 different characters and 20 images per character. Each image is 28 $\times$ 28 gray-scale. We take 1200, 100, 323 classes for training, validation and test, respectively. \textbf{3) \textit{mini}ImageNet.} It is a subset of ImageNet \cite{russakovsky2015imagenet} and consists of 100 classes, 600 images per class with size 84 $\times$ 84. we take 64 classes for training, 16 for validation	and 20 for test, respectively. \textbf{4) CelebA.} CelebA consists of 202,599 face images with 10,177 number of identities. It has been used in the 5-way few-shot recognition task.

\noindent\textbf{Evaluation metrics.} During meta-test, we use the classes in the test set to generate 1000 tasks and compute the mean accuracy and 95\% confidence interval on these tasks. 

\noindent\textbf{Implementation Details.} We adopt the high-level feature reconstruction objective for toy dataset, \textit{mini}-ImageNet and CelebA dataset.  The backbone, variational posterior network $q_\phi(\mathbf{z}|\mathbf{x},\mathcal{T}_t)$ and the high-level feature extractor  (\ie, SimCLR \cite{chen2020simple}) are same as \cite{lee2021metagmvae} for fair comparisons (\ie, 4-layer CNN for Omniglot and 5-layer CNN for others). For the generative network $p_\theta(\mathbf{x}|\mathbf{z},\mathbf{e})$, we concatenate $\mathbf{z}$ and $\mathbf{e}$ in the last dimension, and it outputs the parameter of Bernoulli distribution for Omniglot and the mean of Gaussian distribution for \textit{mini}ImageNet and CelebA. The causal function $h$ is defined as described in section \ref{sec:method-Causal}. There are no other parameters in $q_\phi(\mathbf{e}|\mathbf{z},\mathbf{x})$. 
The number of iterations for causal-EM steps of all experiment is 10. The hyper-parameters $\gamma, \lambda_1$ and $\lambda_2$ are chosen based on the validation class accuracy. We train all models for 60,000 iterations using Adam \cite{kingma2015adam}.

\begin{table}[t]
	\centering
	\caption{Accuracy results on CelebA with 5-way, $S$-shot identity recognition. All the values are from \cite{khodadadeh2021unsupervised}, except for ours and $^\dagger$ that we reproduce.}
	\label{table:celeba}
	\scalebox{0.8}{
		\begin{tabular}{lcc}
			\toprule
			Algorithm                       &     $S=1$      &     $S=5$      \\ \midrule
			Training from scratch           &     34.69      &     56.50      \\
			CACTUs                          &     41.42      &     62.71      \\
			UMTRA                           &     39.30      &     60.44      \\
			LASIUM-RO-GAN-MAML              &     43.88      &     66.98      \\
			LASIUM-RO-VAE-MAML              &     41.25      &     58.22      \\
			LASIUM-RO-GAN-ProtoNets         &     44.39      &     60.83      \\
			LASIUM-RO-VAE-ProtoNets         &     43.22      &     61.12      \\
			Meta-GMVAE$^\dagger$            &     58.05      &     71.95      \\
			IFSL$^\dagger$                  &     57.98      &     72.09      \\ \midrule
			CMVAE (\textit{Ours})           & \textbf{61.04} & \textbf{74.18} \\ \midrule
			MAML (\textit{Supervised})      &     85.46      &     94.98      \\
			ProtoNets (\textit{Supervised}) &     84.17      &     90.84      \\ \bottomrule
		\end{tabular}   
	}
\end{table}

\subsection{Baselines} 
We compare the following unsupervised meta-learning baselines with our approach.  \textbf{CACTUs} \cite{hsu2018unsupervised} extract features by ACAI \cite{berthelot2019understanding}, BiGAN \cite{Donahue2017adversarial}, and Deep-
Cluster \cite{caron2018deep} and then train MAML or ProtoNets. \textbf{UMTRA} \cite{khodadadeh2018unsupervised} generates training tasks by random sampling and augmentation for unsupervised meta-training. \textbf{Meta-GMVAE} \cite{lee2021metagmvae} learns a set-level latent representation by EM algorithm. \textbf{LASIUMs} \cite{khodadadeh2021unsupervised} creates synthetic training data by adding Noise, Random Out-of-class samples, and then train MAML or ProtoNets. \textbf{IFSL} \cite{yue2020inter} is a supervised method. We reimplement it by using backdoor adjustments in Meta-GMVAE. Furthermore, we compare the classic supervised methods  \textbf{MAML} \cite {finn2017model}, \textbf{ProtoNets} \cite{snellPrototypicalNetworksFewshot2017} to indicate the gap between the supervised and unsupervised methods.

\subsection{Results}

\textbf{Toy dataset.}
The 2-way 4-shot classification results in the toy dataset are 78.51 $\pm$ 0.36 for Meta-GMVAE and 93.08 $\pm$ 0.32 for our CMVAE. Since Meta-GMVAE does not take into account the context-bias, its performance is not impressive. While our CMVAE notices the existence of context-bias, the about 15\% improvement on the biased toy dataset demonstrates that it offers the ability to alleviate the context-bias.
\textbf{Natural dataset.}
Table \ref{tab:rest_o_m}  reports the results of few-shot image classification for Omniglot and \textit{mini}ImageNet benchmarks. Table \ref{table:celeba} shows the results of 5-way few-shot identity recognition on CelebA. We can observe that our method outperforms state-of-the-art methods, except for the UMTRA on the 20-shot 5-shot classification in Omniglot. Our CMVAE even outperforms 5-way 1-shot classification supervised MAML in Omniglot. It is noticed that, for challenging dataset \eg, \textit{mini}ImageNet, our method outperforms Meta-GMVAE by more than about 2.5\% average. This shows that 1) Our meta-learning network can capture the causal multi-modal distribution. 2) The causality is a more reliable in the natural images. 3) With causally independent codes and the adjustment for intervention, the confounding effect of meta-knowledge are removed.

\begin{table}[]
	\centering
	\caption{ Results of 5-way 1-shot classification on Omniglot, \textit{mini}ImageNet and CelebA with different settings. We show the impact of choosing hyper parameters on test accuracies. In the Default, the causal function is non-linear, $\lambda_1=1$, $\lambda_2=10^{-4}$, and $\gamma=1$. } 
	\label{tab:ablation}
	\scalebox{0.76}{
		\begin{tabular}{lccc}
			\toprule
			&         Omniglot          &   \textit{mini}ImageNet   &          CelebA           \\ \midrule
			Default            & \textbf{95.11 $\pm$ 0.47} &     43.91 $\pm$ 0.74      &     59.93 $\pm$ 0.95      \\ \midrule
			Linear              &     89.26 $\pm$ 0.56      &     42.68 $\pm$ 0.72      &     51.28 $\pm$ 0.91      \\ \midrule
			$\lambda_1=10^{-1}$ &     94.46 $\pm$ 0.49      &     43.06 $\pm$ 0.75      &     59.84 $\pm$ 0.95      \\
			$\lambda_1=10^5$    &     94.42 $\pm$ 0.48      &     42.11 $\pm$ 0.74      &     59.72 $\pm$ 0.88      \\
			$\lambda_1=10^{10}$ &     91.28 $\pm$ 0.61      &     41.27 $\pm$ 0.70      &     50.92 $\pm$ 0.92      \\
			$\lambda_2=10^{-2}$ &     94.91 $\pm$ 0.50      &     42.88 $\pm$ 0.75      &     59.29 $\pm$ 0.93      \\
			$\lambda_2=10^{-3}$ &     94.95 $\pm$ 0.48      &     43.05 $\pm$ 0.75      &     59.27 $\pm$ 0.94      \\
			$\lambda_2=10^{-5}$ &     93.58 $\pm$ 0.49      &     42.66 $\pm$ 0.72      &     59.59 $\pm$ 0.95      \\ \midrule
			$\gamma^2=s_k^2$    &     90.34 $\pm$ 0.65      &     42.55 $\pm$ 0.75      &     54.25 $\pm$ 0.97      \\
			$\gamma^2=0.5$      &     94.76 $\pm$ 0.48      &     43.48 $\pm$ 0.74      &     60.25 $\pm$ 0.94      \\
			$\gamma^2=0.9$      &     92.40 $\pm$ 0.57      &     43.46 $\pm$ 0.74      & \textbf{61.04 $\pm$ 0.94} \\
			$\gamma^2=5$        &     92.52 $\pm$ 0.54      & \textbf{44.27 $\pm$ 0.76} &     59.04 $\pm$ 0.92      \\
			$\gamma^2=10$       &     58.29 $\pm$ 1.07      &     44.11 $\pm$ 0.75      &     59.04 $\pm$ 0.95      \\ \bottomrule
		\end{tabular}
	}
\end{table}

\noindent\textbf{Visualization.}
To better understand how CMVAE learns in the supervised meta-test stage, we  visualize the real instances and ones generated by $p_\theta(\mathbf{x}|\mathbf{z},\mathbf{e})$ in Figure \ref{fig:res_a}, \ref{fig:res_b}, where each row represents each modality. We can observe that 1) The distinction between real samples and generated samples reveals how well our generative ability for network $p(x|z,h)$ from output distribution. 2) Our CMVAE can capture the similar visual structure in each modality and make it as a class-concept in the meta-test stage.

\subsection{Ablation Study}

\noindent\textbf{Counterfactual samples. \label{sec:count}} 
To further demonstrate the effectiveness of the causality learned by CMVAE, we plot the DAG structure after obtaining $\mathbf{A}$ based on $h$ in Figure \ref{fig:dags}. The nodes are a collection of dimensions of latent codes \ie,  $\mathbf{V}=\{\mathbf{z}_0, \cdots, \mathbf{z}_{63} \}$, and the edges represent cause-to-effect. Note that all the nodes are codes with semantics of interest. We can  discover that $\mathbf{z}_{1},\cdots, \mathbf{z}_{59}$ are the causes.

Figure \ref{fig:dag_b} shows the intervention propagation. Because intervening causes will change the effects while intervening effects will not change the causes, the image will change more massively when intervening causes.  Although we do not know which parts of the image these causes are responsible for generating, they are the most relevant to image generation. To this end, we generate counterfactual samples by intervening the causes and the effects, respectively, with the same amount (\eg, 7 causes or 7 effects) and intervention value (\eg, fixed to 0). Figure \ref{fig:res_c}, \ref{fig:res_d} show the visual results. Comparing them, we conclude as follows: 1) Intervention on the causes from the DAG results in larger changes. Since the intervention can propagate from causes to effects, the DAG learned by our CMVAE is reliable. 2) The causes are the most relevant to the images though we do not know what they means in complex real-world scenes.

\noindent\textbf{DAG type.} We compare the performance of CMVAE with regard to the DAG type, \ie, when the DAG function $h$ is linear or non-linear. The results are shown in the Rows 1-2 of Table \ref{tab:ablation}. We can observe that performances get worse when the function $h$ is linear, which is in line with the common sense that the cause-to-effect is not a simple linear but a complex non-linear relation in the natural images. 

\noindent \textbf{Influence of $\lambda_1,\lambda_2$.} The hyper parameters $\lambda_1,\lambda_2$ control the ``DAGness". The larger $\lambda_1$ and $\lambda_2$, the more strongly causal relations are enforced. Rows 3-8 of Table \ref{tab:ablation} show that the setting when $\lambda_1=1,\lambda_2=10^{-4}$ outperforms other settings. This is because in the real-world images, factors with semantics are unknown and uncountable. The weak constraints  can avoid overfitting the causal relations.

\noindent \textbf{Effects of  $\gamma$.} The value of hype parameter $\gamma$ controls the influence of causal regularization on modalities. We tuned this parameter using the validation classes with the following values: $[s_k^2, 0.5, 0.9, 1.0, 5, 10]$ where  $ s_k^2 = \sum_i(\frac{w_{ik}}{\sum_iw_{ik}})^2 $ for the meta-training and $ s_k^2 = \sum_{s} (\frac{\tilde{\omega}_{sk}}{\sum_s\tilde{\omega}_{sk}+\sum_q\tilde{\omega}_{qk}})^2 + \sum_{q} (\frac{\tilde{\omega}_{qk}}{\sum_s\tilde{\omega}_{sk}+\sum_q\tilde{\omega}_{qk}})^2$ for the meta-test based on the causal EM algorithm, and select the best $\gamma$ corresponding to the best average 5-way 1-shot accuracy over meta-validation data  for inference over the meta-test data. The last 5 rows of Table \ref{tab:ablation} shows the test class accuracies with respect to  different values of $\gamma$. Though $\gamma^2=1,\gamma^2=5,\gamma^2=0.9$ provide the best results for Omniglot, \textit{mini}ImageNet and  CelebA, which shows that the causal regularization needs to satisfy for different datasets, the default value already  outperforms SOTA and it is user-friendly in practice.


\noindent\textbf{Time complexity.} Compared with the original EM, the inference of causal-EM comes with more time cost, as matrix operations (\ie, inversion) have cubic time complexity. Table \ref{tab:time} reports that causal-EM costs about 10\% more time, which is acceptable compared to the better accuracy.

\begin{table}[]
	\centering
	\caption{Time (s) cost  over 10000 20-way tasks on Omniglot during the meta-test stage. Inverse: Matrix inversion.}  
	\label{tab:time}
	\scalebox{0.76}{
		\begin{tabular}{llll}
			\toprule
			& EM & Inverse   & Causal EM \\ \midrule
			1-shot &   129.59         &   139.45 $_{(+7.6\%)}$       &   145.31 $_{(+12.1\%)}$   \\
			5-shot &   143.36        &   150.52 $_{(+5.0\%)}$       &   156.46 $_{(+9.1\%)}$   \\ \bottomrule
		\end{tabular} 
	}
\end{table}

\section{Conclusion}
The context-bias arises when the priors cause spurious corrections between inputs and predictions in unsupervised meta-learning. In this work, we offer an adjustment formulation that performs intervention on inputs to achieve bias-removal. We also develop CMVAE that carries out classification in  causal latent space. Extensive experiments demonstrate that our approach has a better generalization ability across different tasks and datasets. CMVAE is also flexible for the  extension to supervised learning. The limitation is that CMVAE may lack identifiability without any additional observation. We leave these questions for future work.

\bibliography{aaai23}

\newpage
\appendix

\section*{Appendix}

	\begin{figure*}[!h]
	\centering
	\begin{subfigure}[t]{.25\textwidth}
		\centering
		\includegraphics[width=\linewidth]{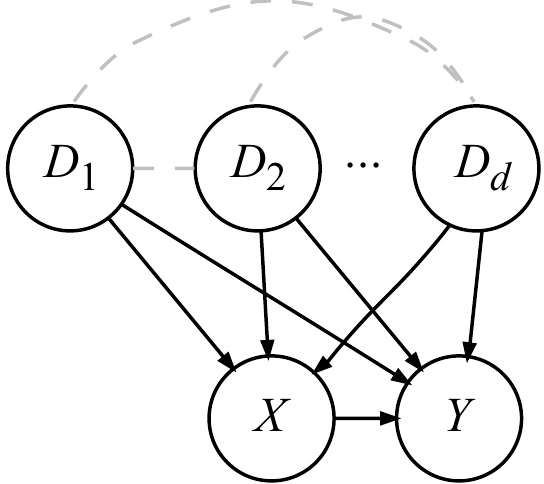}
		\caption{SCM}
		\label{fig:scm}
	\end{subfigure}
	\hfill
	\begin{subfigure}[t]{.55\textwidth}
		\centering
		\includegraphics[width=\linewidth]{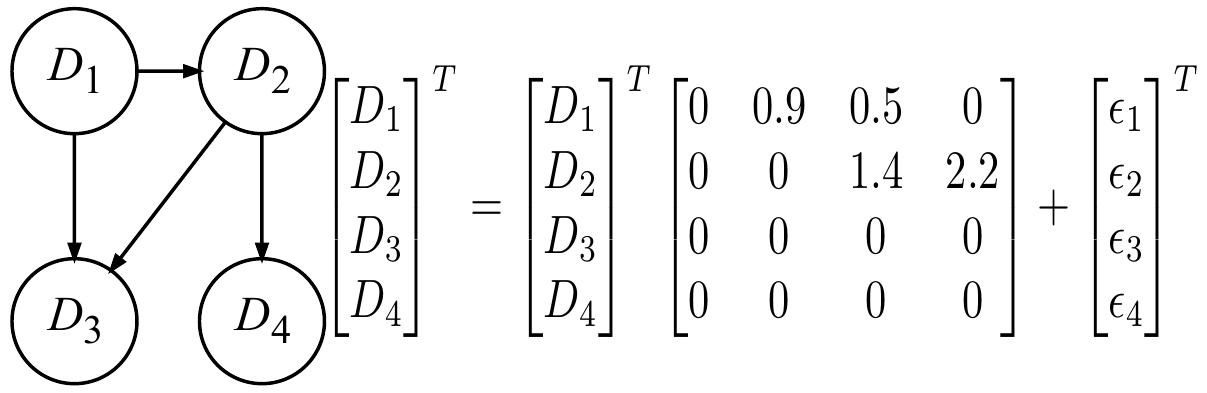}
		\caption{A linear SEM with 4 nodes}
		\label{fig:sem}
	\end{subfigure}	
	\caption{(a) The Structural Causal Model (SCM). Causalities need to be learned  (dashed lines)  (b) A Structural Equation Model (SEM). [Left]:  DAG with 4 nodes. [Right]: A linear equation for Gaussian SEM with noise $\epsilon \sim \mathcal{N}(0, \bm{I})$.}
	\label{fig:ifsl__}
\end{figure*}

\section{Impacts} 
\subsection{Impacts of Unsupervised Meta-Learning}
Though unsupervised meta-learning may not attract much attention now, we argue that it is a promising direction. Supervised meta-learning requires a large labeled dataset during the meta-training phase, which is a limitation in practice. However, unsupervised meta-learning learns to learn with easily obtainable unlabeled datasets in meta-training and only requires few labeled data in meta-test, which is ``actuall'' few-shot learning. Furthermore, in more simple datasets (\ie, Omniglot) 5-way 1-shot task, our work even outperforms supervised MAML. And in more complex datasets (\ie, miniImageNet) 5-way 1-shot task, our work is 2.6\% lower than MAML, where the gap is not too large.

\subsection{Potential Societal Impacts}
Briefly, Our method using the latent variables could be used to alter certain image semantic aspects, and then create fake images with the intent to deceive the system and spread misinformation. Additionally, for causal inference practitioners may over-rely on the claim with few assumptions, becoming less rigorous when considering necessary assumptions such as identifiability. On the other hand, it could have a clear positive social impact, if CMVAE or other unsupervised meta-learning methods become usable and prevalent in application areas such as epidemiology where collecting labeled data is very expensive. CMVAE may also motivate researchers to investigate causal inference, which is a promising area for machine learning.

\section{Basic Causal Properties} \label{app:principle}
\paragraph{Common Cause Principle.} \cite{reichenbach1956direction, scholkopfCausalRepresentationLearning2021} If two observables X and
Y are statistically dependent, then there exists
a variable Z that causally influences both and
explains all the dependence in the sense of making
them independent when conditioned on Z.

\paragraph{Independent Causal Mechanism Principle.} \cite{scholkopfCausalRepresentationLearning2021} The causal generative process of a system’s variables is composed of autonomous modules that do not inform or influence each other. In the probabilistic case, this means that the conditional distribution of each variable given its causes (i.e., its mechanism) does not inform or influence the other mechanisms.

\paragraph{SCM.} 
To describe the relevant concepts and how they interact with each other, we abstract the problem into an SCM in Figure \ref{fig:scm}. In the SCM, $D \to X$ means that the priors D determine where the object appears in an image XX, e.g., in the main paper, the context priors in training images of Figure 1 put the bird object in the sky. $D \to Y$ denotes that the priors D affect the predictions $Y$, e.g. the wing and sky priors lead to the bird prediction. $X \to Y$ is the regular classification process. From the SCM, we observe that $D$ are confounders and cause spurious correlation from $X$ to $Y$.

\paragraph{SEM.}  The causal relationships between the variables $D$ can be estimated via SEM, represented by a weighted DAG. Figure \ref{fig:sem} shows a linear-Gaussian SEM. In this paper, one of the goal is to estimated the weighted DAG.

The function $h_i(u_1,...,u_d)$ does not depend on $u_k$ if $D_k \notin PA(i)$. $h_i$ can show the dependence among variables. For example, given a 4-node DAG where the nodes are $D = \{D_1, D_2, D_3, D_4\}$ and the edges are $\{D_1 \to D_2 \leftarrow D_3, D_4\}$, we have $PA(2) = \{D_1, D_3\}$. Since $D_2$ is not depended on $\{D_2, D_4\}$, the function $h_2$ should be constant for all $u_2, u_4 \in R$ where $D_2=u_2, D_4=u_4$. In other words, if a learned $h_2$ satisfies the above property, the dependence and parents of $D_2$ will be known. Ditto for other $h_i$, and the DAG will be known.

\section{Theoretical Analysis}

\subsection{Derivation for Bias Removal} 
We offer a theoretical analysis based on the SCM in Figure 2. Before the derivation, we first formally introduce two concepts.

\textbf{Definition 1} (Block \cite{glymour2016causal}) A set $D$ of nodes is said to block a path $p$ if either 1) $p$
contains at least one arrow-emitting node that is in $D$, or 2) $p$ contains at least one
collision node that is outside $D$ and has no descendant in $D$. 

\textbf{Definition 2} (Admissible sets \cite{pearl2000models}) A set $D$ is admissible (or "sufficient") for adjustment if two conditions hold: 1). No element of $D$ is a descendant of $X$. 2). The elements of $D$ "block" all "back-door" paths from $X$ to $Y$, namely all paths that end with an arrow pointing to X.

In the SCM, there exists "back-door" paths $P$: $X\leftarrow D_1 \to Y,...,X\leftarrow D_d \to Y$ that carry spurious associations from $X$ to $Y$. Blocking the paths $P$ ensures that the measured association between $X$ and $Y$ is purely causative. Meanwhile, the set $D=\{D_1,...,D_d\}$ is an admissible set and sufficient for adjustment. 

\begin{figure}
	\centering
	\includegraphics[width=0.97\linewidth]{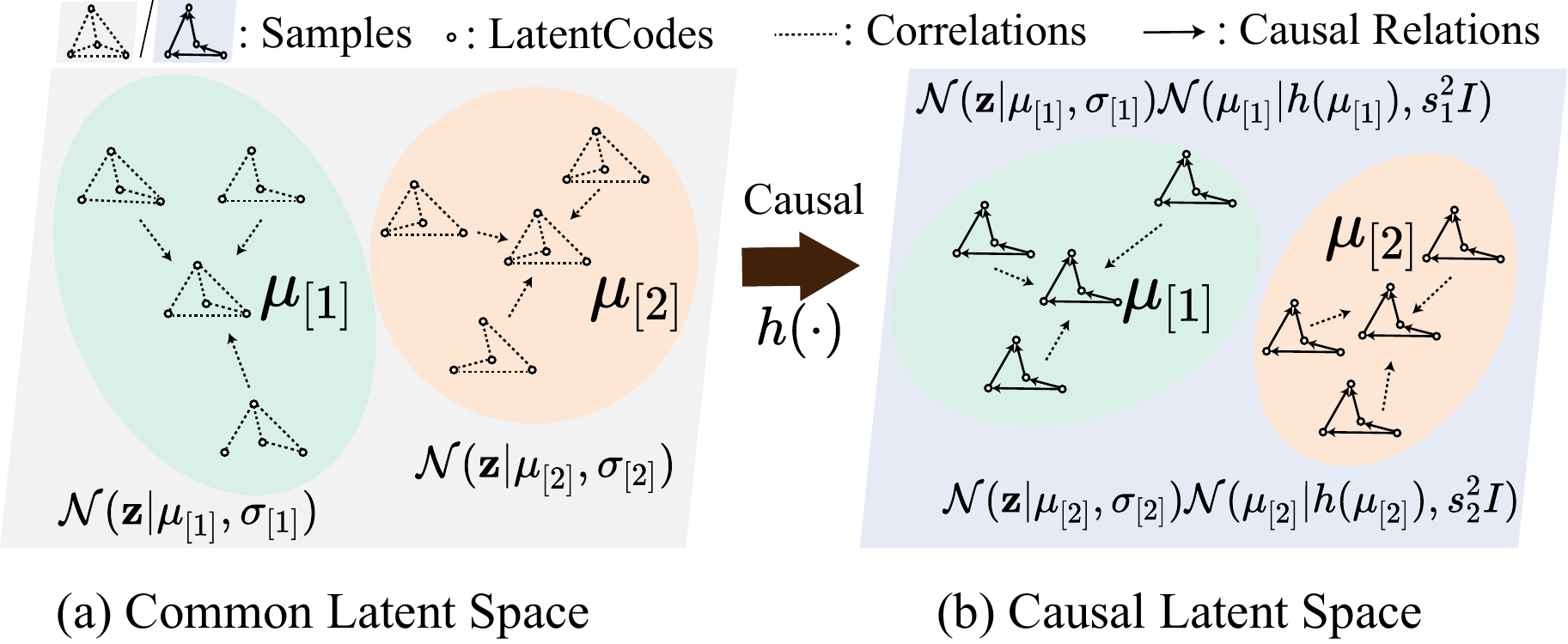}
	\caption{ CMVAE projects the latent codes (left) into the Causal space (right) and performs causal-EM algorithm to get the causal multi-modal prior. During training, the common latent space gradually turns into the causal latent space with the novel loss function.}	
	\label{fig:app_cmvae}
\end{figure}

We consider the binary problem for convenience. Formally, the average risk difference in stratum $\{u_1,...,u_d\}$ of $D$ is:
\begin{align}
	\sum_{u_1,...,u_d}&( P(Y=1|X=1,D_1=u_1,...,D_d=u_d) \notag \\ 
	& - P(Y=1|X=0,D_1=u_1,...,D_d=u_d)) \cdot  \notag \\
	&   P(D_1=u_1,...,D_d=u_d)
\end{align}

The risk difference focuses on effect of $X$. It is considered one estimate of the true causal relationship without bias.
One the other hand, based on the adjustment formula, the causal effect is 
\begin{equation}
	P(Y=1|do(X=1))-P(Y=1|do(X=0))
\end{equation} 
We then derive to prove that the above two equations are equivalent. Firstly, according to the the law of total probability, we have:
\begin{align}
	& P(Y=1|do(X=1))=  \notag  \\ 
	& \sum_{u_1,...,u_d} P(Y=1|do(X=1),D_1=u_1,...,D_d=u_d) \cdot \notag \\
	& \qquad \quad P(D_1=u_1,...,D_d=u_d|do(X=1))
\end{align}

\begin{figure}
	\centering
	\includegraphics[width=0.45\linewidth]{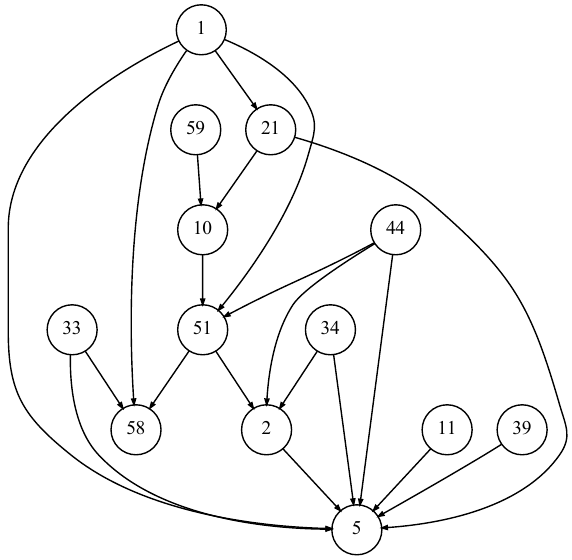}
	\caption{DAG on Omniglot by the learned $\mathbf{A}$. Each node represents each dimension of $\mathbf{z}$. Other nodes are not shown because they are independent and have no cause-to-effect relationship}	
	\label{fig:app_dag_full}
\end{figure}

Since $D_1,...,D_d$ block all backdoor paths $P$, the only connection from $X$ to $Y$ is causal relation. Then, we can remove the $do()$ operator in the factor for $Y$ in the first part and get:
\begin{align}
	& P(Y=1|do(X=1),D_1=u_1,...,D_d=u_d)  \notag \\
	& = P(Y=1|X=1,D_1=u_1,...,D_d=u_d)
\end{align}
Futhermore, there is no path from $X$ to $D_1,..D_d$ since we intervene on $X$. Then we can remove the $do(X=1)$ in the second part:
\begin{align}
	& P(D_1=u_1,...,D_d=u_d|do(X=1))  \notag \\ 
	& =P(D_1=u_1,...,D_d=u_d)
\end{align}
Combining them, we have:
\begin{align}
	P&(Y=1|do(X=1))= \notag \\
	&\sum_{u_1,...,u_d} P(Y=1|X=1,D_1=u_1,...,D_d=u_d) \cdot \notag \\ 
	& \quad \qquad P(D_1=u_1,...,D_d=u_d)
\end{align}

\begin{table*}[t]
	\centering
	\caption{Set-level variational posterior network $q_\phi(\mathbf{z} | \mathbf{x}, \mathcal{T}_t)$ used for Omniglot dataset.}  
	\label{tab:app_set_omn}
	\begin{tabular}{ll}
		\toprule
		Output Size                 & Layers                                                                                           \\ \midrule
		1 $\times$ 28 $\times$ 28   & Input Images                                                                                     \\
		64  $\times$ 14 $\times$ 14 & conv2d(3 $\times$ 3, stride 1, padding 1), BatchNorm2D, ReLU, Maxpool(2 $\times$ 2, stride 2)    \\
		64 $\times$ 7 $\times$ 7    & conv2d(3 $\times$ 3, stride 1, padding 1), BatchNorm2D, ReLU, Maxpool(2 $\times$ 2, stride 2)    \\
		64 $\times$ 4 $\times$ 4    & conv2d(3 $\times$ 3, stride 1, padding 1), BatchNorm2D, ReLU, Maxpool(2 $\times$ 2, stride 2)    \\
		64 $\times$ 2 $\times$ 2    & conv2d(3 $\times$ 3, stride 1, padding 1), BatchNorm2D, ReLU, Maxpool(2 $\times$ 2, stride 2)    \\
		256                         & Flatten                                                                                          \\
		256                         & TransformerEncoder($d_\text{model}$ = 256, $d_\text{ff}$ = 256, $h$ = 4, ELU, LayerNorm = False) \\
		256                         & TransformerEncoder($d_\text{model}$ = 256, $d_\text{ff}$ = 256, $h$ = 4, ELU, LayerNorm = False) \\
		64 $\times$ 2               & Linear(256, 64 $\times$ 2) \\ \bottomrule
	\end{tabular}
\end{table*}

\begin{algorithm}[t]
	\caption{Causal Meta-test}
	\label{alg:meta-test}
	\begin{algorithmic}
		\STATE {\bfseries Input:} A meta test task $\mathcal{T}$, causal-EM steps \texttt{step}.
		\STATE Draw $\mathbf{z} \sim q_\phi(\mathbf{z}|\mathbf{x}, \mathcal{T})$ in Eq. 9
		\STATE Initialize $\bm{\mu}_{[k]}$
		\STATE Compute $\psi^*$ in Eq. 19 with \texttt{step} causal-EM iteratively 
		\STATE Compute $p(\mathbf{y}_q|\mathbf{x}_q, \mathcal{T})$ in Eq. 17
		\STATE {\bfseries Output:}	Query predictions $\hat{\mathbf{y}}_q$ computed by Eq. 18			
	\end{algorithmic}
\end{algorithm}

Similarity,
\begin{align}
	P&(Y=1|do(X=0))= \notag \\
	&\sum_{u_1,...,u_d} P(Y=1|X=0,D_1=u_1,...,D_d=u_d)\cdot \notag \\ 
	&  \quad \qquad P(D_1=u_1,...,D_d=u_d)
\end{align}

Then $P(Y=1|do(X=1))-P(Y=1|do(X=0)$ equals the average risk difference.

In general, for multi-classification, given the admissible set, all factors on the right hand side of the equation are estimable from the observed data, the causal effect can likewise be estimated from such data without bias.

\subsection{Proof of Proposition 3.1 } \label{app:prop1}
Assume $\mathbf{z}_{[i]} \in \mathbb{R}^{1\times d} $ and $w_i \in \mathbb{R}$ is the $i$th element of $\mathbf{Z}$ and $\bm{w}$. Then, we have
\begin{equation}
	\overline{\mathbf{z}} = \sum{(w_i\mathbf{z}_{[i]})} \sim \mathcal{N}(\sum{w_ih(\mathbf{z}_{[i]})}, \sum{{w_i}^2} \bm{I}) \label{eq:taylor1}
\end{equation}
Taking first-order Taylor approximation $h(\mathbf{z})=h(\bm{0})+\mathbf{z}h'(\bm{0}) + R_1(\mathbf{z})$, where $ R_1(\mathbf{z})=\mathbf{z}^2\frac{h''(\bm{\xi} )}{2} $. $h'(0)$ is an amendatory approximation of the DAG matrix, and $R_1(\mathbf{z})$ affects the edge weights only, then
\begin{align}
	& \sum{w_ih(\mathbf{z}_{[i]}i)} \notag \\
	&= \sum{[w_i h(\bm{0})+w_i \mathbf{z}_{[i]}h'(\bm{0})  + w_i R_1(\mathbf{z}_{[i]})]} \notag \\  
	&= h(\bm{0}) +(\sum{ w_i \mathbf{z}_{[i]}}) h'(\bm{0}) +  R_1(\sum{w_i\mathbf{z}_{[i]})} \notag \\
	& \quad \quad \quad \quad  + \sum{ w_i R_1(\mathbf{z}_{[i]})}  - R_1(\sum{w_i\mathbf{z}_{[i]}}) \notag \\ 
	&= h(\sum{ w_i \mathbf{z}_{[i]}}) +  \sum{ w_i R_1(\mathbf{z}_{[i]})}  - R_1(\sum{w_i\mathbf{z}_{[i]}})       \label{eq:taylor2}                  
\end{align}

\begin{table}[]
	\centering
	\caption{Generative Network $p_\theta(\mathbf{x}|\mathbf{z},\mathbf{e})$  for Omniglot dataset.}  
	\label{tab:app_gen_omn}
	
	\begin{tabular}{ll}
		\hline
		Output Size                & Layers                                                                                                    \\ \hline
		64 $\times$ 2              & Latent code                                                                                               \\
		256                        & Linear(64,256), ELU                                                                                       \\
		256                        & Linear(256,256), ELU                                                                                      \\
		256                        & Linear(256,256), ELU                                                                                      \\
		64 $\times$ 2 $\times$ 2   & Unflatten                                                                                                 \\ \hline
		64 $\times$ 4 $\times$ 4   & \begin{tabular}[c]{@{}l@{}}deconv2d(4 $\times$ 4, stride 2, padding 1), \\ BatchNorm2D, ReLU\end{tabular} \\ \hline
		64 $\times$ 7 $\times$ 7   & \begin{tabular}[c]{@{}l@{}}deconv2d(3 $\times$ 3, stride 2, padding 1), \\ BatchNorm2D, ReLU\end{tabular} \\ \hline
		64 $\times$ 14 $\times$ 14 & \begin{tabular}[c]{@{}l@{}}deconv2d(4 $\times$ 4, stride 2, padding 1),\\ BatchNorm2D, ReLU\end{tabular}  \\ \hline
		1 $\times$ 28 $\times$ 28  & \begin{tabular}[c]{@{}l@{}}deconv2d(4 $\times$ 4, stride 2, padding 1), \\ Sigmoid\end{tabular}           \\ \hline
	\end{tabular}
\end{table}

Compared the last two terms:
\begin{align}
	& \lim_{\mathbf{z}_{[i]} \to \bm{0}}  \frac{  \sum{ w_i R_1(\mathbf{z}_{[i]})}  }{ R_1(\sum{w_i\mathbf{z}_{[i]}}) } =  \lim_{\mathbf{z}_{[i]} \to \bm{0}} \frac{   \sum{w_i{\mathbf{z}_{[i]}}^2  } h''(\bm{\xi}) }{  {(\sum{w_i\mathbf{z}_{[i]})}^2 } h''(\bm{\xi}' ) } \notag \\ 
	& =  \frac{ \sum{w_i} h''(\bm{\xi})}{ (\sum{w_i})^2 h''(\bm{\xi}') } = \frac{ h''(\bm{\xi})}{ h''(\bm{\xi}')}	     \label{eq:taylor3}    	                        
\end{align}
where some $ \bm{\xi}_j \in (0, \min_i{(\mathbf{z}_{[i]j}))} , \bm{\xi}_j' \in (0, \sum_i{w_i\mathbf{z}_{[i]j}})$. Combined with Eq. \ref{eq:taylor1}-\ref{eq:taylor3}, we have
\begin{align}
	\overline{\mathbf{z}} &\sim \mathcal{N}( h(\overline{\mathbf{z}}) + (\frac{ h''(\bm{\xi})}{ h''(\bm{\xi}')}-\bm{I})R_1(\overline{\mathbf{z}}), \bm{w}^T\bm{w} \bm{I}) \notag \\
	&\sim \mathcal{N}( h(\overline{\mathbf{z}}), \bm{w}^T\bm{w} \bm{I} ) +  (\frac{ h''(\bm{\xi})}{ h''(\bm{\xi}')}-\bm{I})R_1(\overline{\mathbf{z}})  
\end{align}
Because  $ (\frac{ h''(\bm{\xi})}{ h''(\bm{\xi}')}-\bm{I})R_1(\overline{\mathbf{z}})$ only changes the causal edge weights, it can be ignored and the DAG structure remains unchanged whenever $h$ is linear or non-linear function:

\begin{table*}[]
	\centering
	\caption{Set-level variational posterior network $q_\phi(\mathbf{z} | \mathbf{x}, \mathcal{T}_t)$ used for \textit{mini}ImageNet and CelebA.}  
	\label{tab:app_set_mini}
	\begin{tabular}{ll}
		\toprule
		Output Size                 & Layers                                                                                           \\ \midrule
		256                         & Flatten                                                                                          \\
		256                         & TransformerEncoder($d_\text{model}$ = 256, $d_\text{ff}$ = 256, $h$ = 4, ELU, LayerNorm = False) \\
		256                         & TransformerEncoder($d_\text{model}$ = 256, $d_\text{ff}$ = 256, $h$ = 4, ELU, LayerNorm = False) \\
		64 $\times$ 2               & Linear(256, 64 $\times$ 2) \\ \bottomrule
	\end{tabular}
	
\end{table*}

\begin{table*}[]
	\centering
	\caption{Feature Extractor for SimCLR on \textit{mini}ImageNet and CelebA.}  
	\label{tab:app_simclr}
	\begin{tabular}{ll}
		\toprule
		Output Size                 & Layers                                                                                           \\ \midrule
		3 $\times$ 84 $\times$ 84   & Input Images                                                                                     \\
		64  $\times$ 42 $\times$ 42 & conv2d(3 $\times$ 3, stride 1, padding 1), BatchNorm2D, ReLU, Maxpool(2 $\times$ 2, stride 2)    \\
		64 $\times$ 21 $\times$ 21    & conv2d(3 $\times$ 3, stride 1, padding 1), BatchNorm2D, ReLU, Maxpool(2 $\times$ 2, stride 2)    \\
		64 $\times$ 10 $\times$ 10    & conv2d(3 $\times$ 3, stride 1, padding 1), BatchNorm2D, ReLU, Maxpool(2 $\times$ 2, stride 2)    \\
		64 $\times$ 5 $\times$ 5    & conv2d(3 $\times$ 3, stride 1, padding 1), BatchNorm2D, ReLU, Maxpool(2 $\times$ 2, stride 2)    \\
		64 $\times$ 2 $\times$ 2    & conv2d(3 $\times$ 3, stride 1, padding 1), BatchNorm2D, ReLU, Maxpool(2 $\times$ 2, stride 2)    \\
		256                         & Flatten                                                                                          \\ \bottomrule
	\end{tabular}
\end{table*}
\begin{table}[]
	\centering
	\caption{Generative Network $p_\theta(\mathbf{x}|\mathbf{z},\mathbf{e})$  for \textit{mini}ImageNet and CelebA.}  
	\label{tab:app_gen_mini}
	\begin{tabular}{ll}
		\toprule
		Output Size                 & Layers                                                                                           \\ \midrule
		64 $\times$ 2  & Latent code \\
		512 &Linear(64, 512), ELU\\
		512 &Linear(512, 512), ELU\\
		256 &Linear(512, 256), ELU\\  \bottomrule
	\end{tabular}
\end{table}
\begin{table}[t]
	\centering
	\caption{Results (way, shot) with 95\% confidence interval on the Omniglot.}  
	\label{tab:app_res_omn}
	\begin{tabular}{lll}
		\toprule
		Omniglot (way, shot) & (5,1)             & (5,5)              \\ \hline
		\rowcolor{gray!20}  CMVAE    & 95.11 $\pm$ 0.47  & 97.14 $\pm$ 0.20   \\ \hline
		Omniglot (way, shot) & (20,1)            & (20,5)             \\ \hline
		\rowcolor{gray!20}  CMVAE    & 82.58  $\pm$ 0.41 & 90.97  $\pm$  0.18 \\ \bottomrule
	\end{tabular}
	
\end{table}

\subsection{Details of Equation 8} \label{app:elbo}
In this section, we describe the Equation 8 in detail. Given a task $\mathcal{T}_t$, we assume there exists task dependent causal multi-modalities $\psi_t^*$, and we want maximize the marginal log-likehood of  $\mathcal{T}_t$:
\begin{small}
	\begin{align}
		& \log p_\theta(\mathcal{T}_t) = \sum\log p_\theta(\mathbf{x}) \notag \\ 
		& =\sum\log \iint  p_\theta(\mathbf{x}|\mathbf{z}, \mathbf{e})p_{\psi_t^*}(\mathbf{z})p(\mathbf{e}|\mathbf{z})\frac{q_\phi(\mathbf{e},\mathbf{z}|\mathbf{x},\mathcal{T}_t)}{q_\phi(\mathbf{e},\mathbf{z}|\mathbf{x},\mathcal{T}_t)}d\mathbf{e}d\mathbf{z} \notag \\
		& = \sum\log \iint p_\theta(\mathbf{x}|\mathbf{z}, \mathbf{e})p_{\psi_t^*}(\mathbf{z})p(\mathbf{e}|\mathbf{z})\frac{q_\phi(\mathbf{z}|\mathbf{x},\mathcal{T}_t)q_\phi(\mathbf{e}|\mathbf{z},\mathbf{x})}{q_\phi(\mathbf{z}|\mathbf{x},\mathcal{T}_t)q_\phi(\mathbf{e}|\mathbf{z},\mathbf{x})}d\mathbf{e}d\mathbf{z} \notag \\
		& = \sum\log \int p_\theta(\mathbf{x}|\mathbf{z}, \mathbf{e})p(\mathbf{e}|\mathbf{z})
		\frac{q_\phi(\mathbf{e}|\mathbf{z},\mathbf{x})}{q_\phi(\mathbf{e}|\mathbf{z},\mathbf{x})}d\mathbf{e}  \notag \\
		& \qquad \qquad \qquad \int
		p_{\psi_t^*}(\mathbf{z}) \frac{q_\phi(\mathbf{z}|\mathbf{x},\mathcal{T}_t)}{q_\phi(\mathbf{z}|\mathbf{x},\mathcal{T}_t)}d\mathbf{z} \notag \\
		& \ge  \sum\mathbb{E}_{q_\phi(\mathbf{z}|\mathbf{x}, \mathcal{T}_t)}[
		\mathbb{E}_{q_\phi(\mathbf{e}|\mathbf{z},\mathbf{x})}[ 
		\log p_\theta(\mathbf{x}|\mathbf{z},\mathbf{e}) - \log \frac{q_\phi(\mathbf{e}|\mathbf{z},\mathbf{x})}{p(\mathbf{e}|\mathbf{z})} ]	\notag \\
		& \quad +\log  p_{\psi_t^{*}}(\mathbf{z})-\log q_\phi(\mathbf{z}|\mathbf{x}, \mathcal{T}_t)] \notag \\
		& = \sum\text{ELBO} , \quad \mathbf{z} \sim q_\phi(\mathbf{z}|\mathbf{x}, \mathcal{T}_t), \quad 
		\mathbf{e}|\mathbf{z} \sim q_\phi(\mathbf{e}|\mathbf{z},\mathbf{x})
	\end{align}
\end{small}

\begin{table}[]
	\centering
	\caption{Results (way, shot) with 95\% confidence interval on the \textit{mini}ImageNet and CelebA.}  
	\label{tab:app_res_mini}
	\begin{tabular}{lcc}
		\hline
		\textit{mini}ImageNe (way, shot) &       (5,1)       &       (5,5)       \\ \hline
		\rowcolor{gray!20}  CMVAE                            & 44.27 $\pm$ 0.76  & 58.95 $\pm$ 0.71  \\ \hline
		\textit{mini}ImageNe (way, shot) &      (5, 20)      &      (5, 50)     \\ \hline
		\rowcolor{gray!20} 	CMVAE                            & 66.25  $\pm$ 0.51 & 70.54  $\pm$ 0.44 \\ \hline
		CelebA  (way, shot)  &       (5,1)       &       (5,5)       \\ \hline
		\rowcolor{gray!20}  CMVAE                        & 61.04 $\pm$ 0.94  & 74.18 $\pm$ 0.67  \\ \hline		                                 
	\end{tabular}
\end{table}

\subsection{Derivations of Causal-EM} \label{app:em}
The MCP can be rewritten as:
\begin{small}
	\begin{align}
		&\sum_i\log p_{\psi_t}(\mathbf{z}_{[i]}) = \sum_i\log \sum_{k}p_{\psi_t}(\mathbf{z} |c)p_{\psi_t}(c) \notag\\ 
		& = 	 \sum_i\log \sum_{k}p(c=k)\mathcal{N}(\mathbf{z}|\bm{\mu}_{[k]}, {\bm{\sigma}}_{[k]}^2\bm{I}) \mathcal{N}(\bm{\mu}_{[k]}|h(\bm{\mu}_{[k]}), s_k^2\bm{I}) \notag \\
		& = \sum_i\log\sum_kp({c}=k|\mathbf{z}_{[i]}, \bm{\mu}_{[k]}, {\bm{\sigma}}_{[k]}^2\bm{I}, s_k^2\bm{I}) \cdot \notag \\ 
		& \qquad \frac{p({c}=k)\mathcal{N}(\mathbf{z}|\bm{\mu}_{[k]}, {\bm{\sigma}}_{[k]}^2\bm{I}) \mathcal{N}(\bm{\mu}_{[k]}|h(\bm{\mu}_{[k]}), s_k^2\bm{I})}{p({c}=k|\mathbf{z}_{[i]}, \bm{\mu}_{[k]}, {\bm{\sigma}}_{[k]}^2\bm{I})} \notag \\
		&\ge \sum_i\sum_k\omega_{ik}\log\frac{\alpha_k\mathcal{N}(\mathbf{z}|\bm{\mu}_{[k]}, {\bm{\sigma}}_{[k]}^2\bm{I}) \mathcal{N}(\bm{\mu}_{[k]}|h(\bm{\mu}_{[k]}), s_k^2\bm{I})}{\omega_{ik}} \notag \\
		& = \mathcal{Q}(\psi_t, \psi_t')
	\end{align}
\end{small}
\begin{figure*}
	\centering
	\begin{subfigure}[t]{.22\textwidth}
		\centering
		\includegraphics[width=\linewidth]{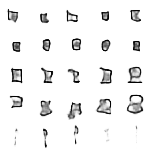}
		\caption{Intervention on $\mathbf{z}_{1},\mathbf{z}_{59}$}
		\label{fig:app_res_a}
	\end{subfigure}
	\hfill
	\begin{subfigure}[t]{.22\textwidth}
		\centering
		\includegraphics[width=\linewidth]{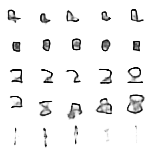}
		\caption{Intervention on $\mathbf{z}_{7},\mathbf{z}_{30}$}
		\label{fig:app_res_b}
	\end{subfigure}		
	\hfill		
	\begin{subfigure}[t]{.22 \textwidth}
		\centering
		\includegraphics[width=\linewidth]{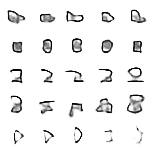}
		\caption{Intervention on $\mathbf{z}_{5},\mathbf{z}_{58}$}
		\label{fig:app_res_c}
	\end{subfigure}
	\hfill
	\begin{subfigure}[t]{.22\textwidth}
		\centering
		\includegraphics[width=\linewidth]{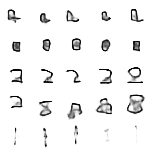}
		\caption{Intervention on $\mathbf{z}_{23},\mathbf{z}_{41}$}
		\label{fig:app_res_d}
	\end{subfigure}
	\medskip
	\centering
	\begin{subfigure}[t]{.22\textwidth}
		\centering
		\includegraphics[width=\linewidth]{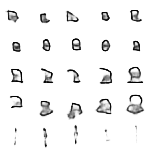}
		\caption{Intervention on $\mathbf{z}_{1},\mathbf{z}_{59}$}
		\label{fig:app_res_e}
	\end{subfigure}
	\hfill
	\begin{subfigure}[t]{.22\textwidth}
		\centering
		\includegraphics[width=\linewidth]{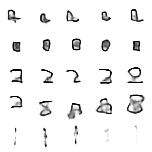}
		\caption{Intervention on $\mathbf{z}_{7},\mathbf{z}_{30}$}
		\label{fig:app_res_f}
	\end{subfigure}		
	\hfill		
	\begin{subfigure}[t]{.22 \textwidth}
		\centering
		\includegraphics[width=\linewidth]{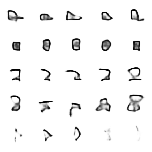}
		\caption{Intervention on $\mathbf{z}_{5},\mathbf{z}_{58}$}
		\label{fig:app_res_g}
	\end{subfigure}
	\hfill
	\begin{subfigure}[t]{.22\textwidth}
		\centering
		\includegraphics[width=\linewidth]{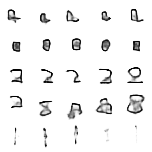}
		\caption{Intervention on $\mathbf{z}_{23},\mathbf{z}_{41}$}
		\label{fig:app_res_h}
	\end{subfigure}
	
	\caption{Counterfactual samples generated. In the top row, the intervened value is 0. In the bottom row, the intervened value is 1. (a, e) The dimensions are two causes.  (c, g) The dimensions are two effects. (b, d, f, h) The dimensions are selected randomly.}  
	\label{fig:app_counter_all}
\end{figure*}

where $\psi_t'$ is values of the previous iteration, $\alpha_k=p({c}=k)$ and $s_k^2=\sum_i(\frac{w_{ik}}{\sum_iw_{ik}})^2$. Here we fix the parameters of $h$ during the whole progress.

\textbf{E-step:} According to Bayes' theorem, $\omega_{ik}$ is:
\begin{align}
	& \omega_{ik}=p({c}=k|\mathbf{z}_{[i]}, \bm{\mu}_{[k]}, {\bm{\sigma}}_{[k]}^2\bm{I}, s_k^2\bm{I}) \notag \\
	& = \frac{\alpha_kp(\mathbf{z}_{[i]}| \bm{\mu}_{[k]}, {\bm{\sigma}}_{[k]}^2\bm{I}, s_k^2\bm{I})}{\sum_k\alpha_kp(\mathbf{z}_{[i]}| \bm{\mu}_{[k]}, {\bm{\sigma}}_{[k]}^2\bm{I}, s_k^2\bm{I})} \notag \\
	& =\frac{\alpha_k \mathcal{N}(\mathbf{z}_{[i]} | \bm{\mu}_{[k]}, \bm{I}) \mathcal{N}(\bm{\mu}_{[k]} | h(\bm{\mu}_{[k]}), s_k^2\bm{I}) }{\sum_k\alpha_k \mathcal{N}(\mathbf{z}_{[i]} | \bm{\mu}_{[k]}, \bm{I}) \mathcal{N}(\bm{\mu}_{[k]} | h(\bm{\mu}_{[k]}), s_k^2\bm{I}) }
\end{align}

Then we have \begin{align}
	& \mathcal{Q}(\psi_t, \psi_t') \notag \\
	& =\sum_i\sum_k\omega_{ik}(\log\alpha_k-\log\omega_{ik}  \notag \\
	& \qquad -  \log\sqrt{2\pi\bm{\sigma}_{[k]}^2} -\frac{
		(\mathbf{z}_{[i]}-\bm{\mu}_{[k]})(\mathbf{z}_{[i]}-\bm{\mu}_{[k]})^T }{  2 \bm{\sigma}_{[k]}^2 } \notag \\
	& \qquad - \log\sqrt{2\pi s_{k}^2} - \frac{
		(\bm{\mu}_{[k]}- h(\bm{\mu}_{[k]}))(\bm{\mu}_{[k]}- h(\bm{\mu}_{[k]})^T }{  2 s_{k}^2 }
	)
\end{align}
\textbf{M-step:}
The derivations of $\alpha_k$ and $\bm{\sigma}_{[k]}$ is the same with common M-step, since the additional term does not contain these parameters. 
\begin{gather}
	\alpha_k = \frac{\sum_{i=1}^M\omega_{ik}}{\sum_{k=1}^K\sum_{i=1}^M\omega_{ik}} \\
	\bm{\sigma}_{[k]} = \frac{\sum_{i=1}^M\omega_{ik}(\mathbf{z}_{[i]} - \bm{\mu}_{[k]})^2}{\sum_{i=1}^M\omega_{ik}}
\end{gather}

For $\bm{\mu}_{[k]}$, we take first-order Taylor approximation $h(\mathbf{z})\approx h(\bm{0})+\mathbf{z}h'(\bm{0})$. $h'(0)$ is an amendatory approximation of the DAG matrix, and others affects the edge weights only. We assume the errors can be eliminated with neural networks. Denote $\epsilon(\cdot)$:
\begin{equation}
	\epsilon(\mathbf{z})=\mathbf{z}-h(\mathbf{z}) \approx  \epsilon(\bm{0})+\mathbf{z}(\bm{I}-h'(\bm{0}))
\end{equation}  
and $\mathbf{b}=\epsilon(\bm{0}), \mathbf{C}=\bm{I}-h'(\bm{0})$, then we have $\epsilon(\mathbf{z})= \mathbf{b}+\mathbf{z}\mathbf{C}$.
\begin{align}
	&\frac{\partial \mathcal{Q}}{\partial \bm{\mu}_{[k]}} = \sum_i\omega_{ik}(\frac{
		\mathbf{z}_i-\bm{\mu}_{[k]} }{\bm{\sigma}_{[k]}^2 }-\frac{\bm{\mu}_{[k]}\mathbf{C} \mathbf{C}^T - \mathbf{C}\mathbf{b}^T }{s_k^2} )  = \bm{0} \notag \\
	\Rightarrow \quad & \bm{\mu}_{[k]} = \frac{\sum_i\mathbf{z}_{[i]}(\bm{I} +\mathbf{C}\mathbf{C}^T(s_k^{-1}\bm{\sigma}_{[k]}\bm{I} )^2 )^{-1}}{\sum_i\omega_{ik}} \notag \\
	& \quad \quad- \mathbf{b}\mathbf{C}^T(s_k^{-1}\bm{\sigma}_{[k]}\bm{I} )^2
\end{align}
Replace $\mathbf{b}, \mathbf{C}$ back with $\epsilon()$,
\begin{align}
	&\mathbf{C}\mathbf{C}^T(s_k^{-1}\bm{\sigma}_{[k]}\bm{I} )^2 = (s_k^{-1}\bm{\sigma}_{[k]}\bm{I}\mathbf{C})(s_k^{-1}\bm{\sigma}_{[k]}\bm{I}\mathbf{C})^T \notag \\
	&\approx (\epsilon(s_k^{-1}\bm{\sigma}_{[k]}\bm{I}) -\epsilon(\mathbf{0}_{[d\times d]}))(\epsilon(s_k^{-1}\bm{\sigma}_{[k]}\bm{I}) -\epsilon(\bm{0}_{[d\times d]}))^T 
\end{align} 
\begin{gather}
	\mathbf{b}\mathbf{C}^T(s_k^{-1}\bm{\sigma}_{[k]}\bm{I} )^2 = s_k^{-2}\bm{\sigma}_{[k]}^2\mathbf{b} \mathbf{C}^T \approx \epsilon(s_k^{-2}\bm{\sigma}_{[k]}^2\epsilon(\bm{0}))-\epsilon(\bm{0})
\end{gather}

\begin{figure*}
	\centering
	\begin{subfigure}[t]{.22\textwidth}
		\centering
		\includegraphics[width=\linewidth]{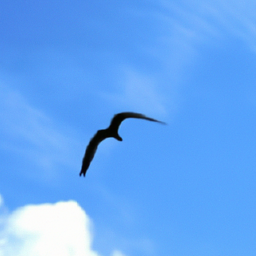}
		\caption{}
		\label{fig:app_toy_1}
	\end{subfigure}
	\hfill
	\begin{subfigure}[t]{.22\textwidth}
		\centering
		\includegraphics[width=\linewidth]{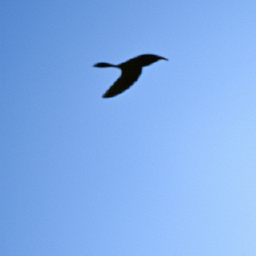}
		\caption{}
		\label{fig:app_toy_2}
	\end{subfigure}		
	\hfill		
	\begin{subfigure}[t]{.22 \textwidth}
		\centering
		\includegraphics[width=\linewidth]{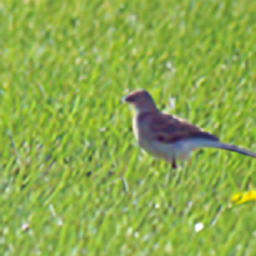}
		\caption{}
		\label{fig:app_toy_3}
	\end{subfigure}
	\hfill
	\begin{subfigure}[t]{.22\textwidth}
		\centering
		\includegraphics[width=\linewidth]{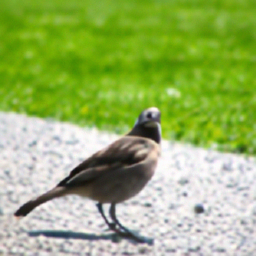}
		\caption{}
		\label{fig:app_toy_4}
	\end{subfigure}
	
	\caption{(a)(b) Toy samples generated by "A bird is flying in the sky". (c)(d) Toy samples generated by "A bird is standing on the ground".  }  
	\label{fig:app_toy_11}
\end{figure*}

\begin{table*}[]
	\centering
	\caption{Results with 95\% confidence interval on the \textit{mini}ImageNet and CelebA.}  
	\label{tab:app_res_each}
	\begin{tabular}{lccccccccc}
		\toprule
		& Method     & AT         & Linear     & Nonlinear  & Vanilla-EM & Causal-EM  & miniImageNet     & CelebA           &  \\ \midrule
		& Baseline   &            &            &            & $\checkmark$          &            & 42.81 $\pm$ 0.70 & 58.05 $\pm$ 0.90 &  \\
		&            & $\checkmark$          & $\checkmark$          &            & $\checkmark$          &            & 42.68 $\pm$ 0.72 & 51.28 $\pm$ 0.91 &  \\
		&            & $\checkmark$          &            & $\checkmark$          & $\checkmark$          &            & 43.48 $\pm$ 0.74 & 60.03 $\pm$ 0.95 &  \\
		& Ours       & $\checkmark$          &            & $\checkmark$          &            & $\checkmark$          & 44.27 $\pm$ 0.76 & 61.04 $\pm$ 0.94 & \\ \bottomrule
	\end{tabular}
	
\end{table*}

Then we can get the approximated closed solution:
\begin{small}
	\begin{align}
		&\bm{\mu}_{[k]} = \notag \\
		& \frac{\sum_i\mathbf{z}_{[i]}(\bm{I} + (\epsilon(s_k^{-1}\bm{\sigma}_{[k]}\bm{I})  -\epsilon(\mathbf{0}_{[d\times d]}))(\epsilon(s_k^{-1}\bm{\sigma}_{[k]}\bm{I}) -\epsilon(\bm{0}_{[d\times d]}))^T )^{-1}}{\sum_i\omega_{ik}} \notag \\
		& - \epsilon(s_k^{-2}\bm{\sigma}_{[k]}^2\epsilon(\bm{0}))+\epsilon(\bm{0}) \label{eq:app_mu}
	\end{align}
\end{small}	
Both in the unsupervised meta-learning and meta-test, we assume  $ \epsilon(\bm{0}) = \bm{0}$ to reduce the complexity of calculation because the errors can be ignored iteratively if $\mathbb{E}f_j(\mathbf{z})=0$. Then Eq \ref{eq:app_mu} can be reduced as:
\begin{equation}
	\bm{\mu}_{[k]} = \frac{\sum_i\mathbf{z}_{[i]}(\bm{I} + \epsilon(s_k^{-1}\bm{\sigma}_{[k]}\bm{I}) \epsilon^T(s_k^{-1}\bm{\sigma}_{[k]}\bm{I})  )^{-1}}{\sum_i\omega_{ik}} \label{eq:app_mu_red}
\end{equation}

\section{Implementation details} \label{app:implementation}
\subsection{High level of CMVAE}
We show high level of CMVAE as Figure \ref{fig:app_cmvae}. CMVAE projects the latent codes (left) into the Causal space (right) and performs causal-EM algorithm to get the causal multi-modal prior. During training, the common latent space gradually turns into the causal latent space.  The Algorithm \ref{alg:meta-test} shows the meta-test stage.

\subsection{Omniglot}
Following Meta-GMVAE \cite{lee2021metagmvae}, we train all models for 60,000 iterations using Adam \cite{kingma2015adam} with learning rate 1e-3. For the 5-way experiments (\ie, K = 5), we set the mini-batch size, the number of datapoints, and Monte Carlo sample size as 4, 200, and 32, respectively. For the 20-way experiments (\ie, K = 20), we set them as 4, 300, and 32. We set the number of causal EM iterations as 10.

\textbf{Network architecture.} The set-level variational posterior network $q_\phi(\mathbf{z} | \mathbf{x}, \mathcal{T}_t)$ and generative Network $p_\theta(\mathbf{x}|\mathbf{z},\mathbf{e})$ are summarized as Table \ref{tab:app_set_omn}, \ref{tab:app_gen_omn}, respectively.

\textbf{95\% Confidence interval.} We provide the standard errors of our model’s performance at 95\% confidence interval over 1000 episodes on the Omniglot dataset in Table \ref{tab:app_res_omn}.

\begin{figure*}
	\centering
	\begin{subfigure}[t]{.22\textwidth}
		\centering
		\includegraphics[width=\linewidth]{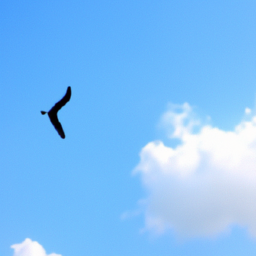}
		\caption{}
	\end{subfigure}
	\hfill
	\begin{subfigure}[t]{.22\textwidth}
		\centering
		\includegraphics[width=\linewidth]{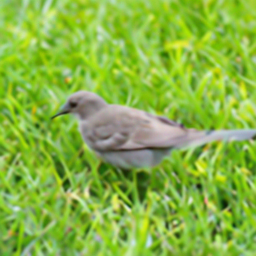}
		\caption{}
		\label{fig:app_toy_21}
	\end{subfigure}		
	\hfill		
	\begin{subfigure}[t]{.22 \textwidth}
		\centering
		\includegraphics[width=\linewidth]{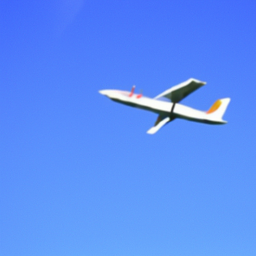}
		\caption{}
		\label{fig:app_toy_31}
	\end{subfigure}
	\hfill
	\begin{subfigure}[t]{.22\textwidth}
		\centering
		\includegraphics[width=\linewidth]{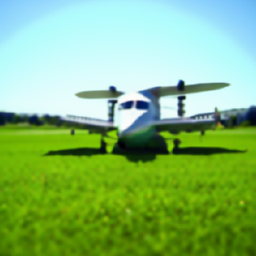}
		\caption{}
		\label{fig:app_toy_41}
	\end{subfigure}
	
	\caption{(a) A toy sample generated by "A bird in the sky". (b) A toy sample generated by "A bird is on the ground" (c) A toy sample generated by "A plane in the sky". (d) A toy sample generated by "A plane is on the ground" }  
	\label{fig:app_toy_12}
\end{figure*}

\subsection{\textit{mini}ImageNet and CelebA}
Since the high-level features for \textit{mini}ImageNet and CelebA are extracted by SimCLR, the settings are also same.  We train all models using Adam \cite{kingma2015adam} with learning rate 1e-4. For the 5/20-way  experiments (\ie, K = 5 or 20), we set the mini-batch size, the number of datapoints, and Monte Carlo sample size as 16, 5, and 256, respectively. We set the number of causal EM iterations as 10.

\textbf{Network architecture.} The SimCLR, set-level variational posterior network $q_\phi(\mathbf{z} | \mathbf{x}, \mathcal{T}_t)$ and generative Network $p_\theta(\mathbf{x}|\mathbf{z},\mathbf{e})$ are summarized as Table \ref{tab:app_simclr}, \ref{tab:app_set_mini}, \ref{tab:app_gen_mini}, respectively.

\textbf{95\% Confidence interval.} We provide the standard errors of our model’s performance at 95\% confidence interval over 1000 episodes on the \textit{mini}ImageNet and CelebA dataset in Table \ref{tab:app_res_mini}.

\section{Additional Study}

\begin{table}[]
	\centering
	\caption{Results with 95\% confidence interval on the biased toy examples.}  
	\label{tab:app_res_others}
	\begin{tabular}{llll}\toprule
		& Method     & 2-way 5-shot     &  \\ \midrule
		& Meta-GMVAE & 78.51 $\pm$ 0.36 &  \\ 
		& Ours       & 93.08 $\pm$ 0.32 &  \\ \bottomrule
	\end{tabular}
\end{table}

\textbf{DAG and Counterfactual Samples.} \label{app:counter}
we show the full DAG in  Figure \ref{fig:app_dag_full}. From Figure \ref{fig:app_dag_full},  we  discover that $\mathbf{z}_{1},\mathbf{z}_{11}, \mathbf{z}_{33},\mathbf{z}_{34},\mathbf{z}_{39},\mathbf{z}_{44}, \mathbf{z}_{59}$ are the causes.
More counterfactual samples are shown in Figure \ref{fig:app_counter_all}.  In the top row, the intervened value is 0. In the bottom row, the intervened value is 1.

\textbf{Results of Each Component.}
Table \ref{tab:app_res_each} reports  the results of the ablation study on each component on 5-way 1-shot classification on miniImageNet and CelebA. 'Baseline' denotes that we do not consider the causality and do not apply the adjusting term. 'AT' denotes that we apply the adjusting term. 'Linear' denotes that we assume the causality relationship between context priors is linear. 'Nonlinear' denotes that we assume the causality relationship between context priors is nonlinear. 'Vanilla-EM' denotes that we calculate the modalities with traditional EM. 'Causal-EM' denotes that we calculate the modalities with causal-EM. We observe that: (1) It is in line with common sense that the cause-to-effect is not a simple linear but a complex nonlinear relationship in the natural images. (2) With nonlinear assumptions, the adjusting term address the context bias. (3) Causal-EM is the solution for maximum causal posterior and can remove the context bias better.

\begin{table}[]
	\centering
	\caption{Results with 95\% confidence interval on the \textit{mini}ImageNet and CelebA.}  
	\label{tab:app_res_other}
	\begin{tabular}{lllll}
		\toprule
		& Method          & miniImageNet     & CelebA           &  \\\midrule
		& DRC  & 43.14 $\pm$ 0.73 & 59.67 $\pm$ 0.91 &  \\
		& STMT & 41.97 $\pm$ 0.65 & 58.92 $\pm$ 0.90 &  \\
		& Ours            & 44.27 $\pm$ 0.76 & 61.04 $\pm$ 0.94 &  \\ \bottomrule
	\end{tabular}
\end{table}

\subsection{Toy Examples}
In this section, we perform two types of toy examples.

\textbf{Intuitively-labeled Toy Example.} This toy example is to further show the learned DAG with an unsupervised manner is meaningful. We build a synthetic bird image dataset. Each image contains 3 concepts (wing, flying, sky). Intuitively, the relationship-label $A_{g}$ is "wing $\leftarrow$ flying $\to$ sky". Figure \ref{fig:app_toy_11} shows some samples. Specially, 1K images with 84 $\times$ 84 size are generated by the Composable Diffusion Models \cite{liu2022compositional} with the text input "A bird is flying in the sky" and "A bird is standing on the ground". During unsupervised training, we set the dimension to 3 and the number of clusters to 1. During test, we calculate the relationship $A$ from the learned DAG, and compute the Structural Hamming distance (SHD). We ran 100 random experiments. For the total 2 edges, the average of SHD is 0.7 in the linear DAG setting, and 0.4 with the nonlinear DAG setting, respectively. (lower is better).

\textbf{Biased Toy Example.}
Here we provide a 2-way biased toy example. We build a synthetic image dataset with two class "bird" and "plane". Specially, 2k images (1K for training and 1K for test) are generated by the Composable Diffusion Models \cite{liu2022compositional}  with the text input "A bird in the sky", "A bird is on the ground", "A plane in the sky" and "A plane is on the ground". Figure \ref{fig:app_toy_12} shows some samples.  As Figure 1 shows, in the tasks, the support sets are biased and the query sets are drawn uniformly at random. During test, the number of clusters is set to 2 and the number of query data is 15.  The results are as shown in Table \ref{tab:app_res_others}. We can observe that our can alleviate the bias with about 15\% improvement.

\subsection{Compared with Other Methods}
For background removal approaches, DRC \cite{yu2021unsupervised} is probabilistic foreground-background modeling by reconciling energy-based prior in a fully unsupervised manner. For probabilistic graphical models (PGMs), STMT \cite{9414435} enriches the dependencies between the random variables to better take into account the spatial context of an image in an unsupervised manner. To apply DRC and STMT in unsupervised meta-learning, we train them in the unlabeled train set to remove the background and extract the foreground and then use standard Meta-GMVAE to perform classification. The results of 5-way 1-shot classification are as shown in Table \ref{tab:app_res_other}. Our method performs best. The reason may be (1) DRC and STMT do not consider the dependence among priors. (2) Meta-GMVAE highly depends on the qualities of DRC and STMT.

\end{document}